%% file: large_am_journal.tex
\documentclass[journal]{IEEEtran}
\usepackage{amssymb,amsmath,epsfig,url,mathabx,tikz}
\usetikzlibrary{positioning,patterns,fit}
\usepackage{booktabs}
\newcommand{\citet}[1]{\cite{#1}}

\begin{document}
\title{Building DNN Acoustic Models for Large Vocabulary Speech Recognition}

\author{Andrew~L.~Maas,
  Peng~Qi,
  Ziang~Xie,
  Awni~Y.~Hannun,
  Christopher~T.~Lengerich,
  Daniel~Jurafsky,
  Andrew~Y.~Ng,%
\thanks{Department of Computer Science, Stanford University, Stanford,
CA, 94305 USA e-mail: amaas@cs.stanford.edu.}
}

%
%


\maketitle

\begin{abstract}
Deep neural networks (DNNs) are now a central component of nearly all
state-of-the-art speech recognition systems. Building neural network
acoustic models requires several design decisions including network
architecture, size, and training loss function. This paper offers an
empirical investigation on which aspects of DNN acoustic model design
are most important for speech recognition system performance.  We
report DNN classifier performance and final speech recognizer word
error rates, and compare DNNs using several metrics to quantify
factors influencing differences in task performance. Our first set of
experiments use the standard Switchboard benchmark corpus, which
contains approximately 300 hours of conversational telephone
speech. We compare standard DNNs to convolutional networks, and
present the first experiments using locally-connected, untied neural
networks for acoustic modeling. We additionally build systems on a
corpus of 2,100 hours of training data by combining the Switchboard
and Fisher corpora. This larger corpus allows us to more thoroughly
examine performance of large DNN models -- with up to ten times more
parameters than those typically used in speech recognition systems.
Our results suggest that a relatively simple DNN architecture and
optimization technique produces strong results. These findings, along
with previous work, help establish a set of best practices for
building DNN hybrid speech recognition systems with maximum likelihood
training. Our experiments in DNN optimization additionally serve as a
case study for training DNNs with discriminative loss functions for
speech tasks, as well as DNN classifiers more generally.
\end{abstract}


\IEEEpeerreviewmaketitle


\input{introduction}

\input{motivation}
\input{questions}

\input{model}
\input{experiment_overfit}

\input{experiment_cnn}

\input{experiment_combined_corpus}
\input{coding_properties}

\input{conclusion}
\ifCLASSOPTIONcaptionsoff
  \newpage
\fi



%



\bibliographystyle{IEEEtran}
\bibliography{audio}

%









\end{document}

%% file: introduction.tex
\section{Introduction}\label{sec:introduction}
\IEEEPARstart{D}{eep} neural network (DNN) acoustic models have driven
tremendous improvements in large vocabulary continuous speech
recognition (LVCSR) in recent years.  Initial research hypothesized
that DNNs work well because of unsupervised pre-training
\cite{Dahl2011a}. However, DNNs with random initialization yield
state-of-the-art LVCSR results for several speech recognition
benchmarks \cite{Hinton2012,Kingsbury2012,Vesely2013}. Instead, it
appears that modern DNN-based systems are quite similar to
long-standing neural network acoustic modeling approaches
\cite{Bourlard93,Hermansky2000,Renals1994}. Modern DNN systems build
on these fundamental approaches but utilize increased computing power,
training corpus size, and function optimization heuristics. This paper
offers a large empirical investigation of DNN performance on two
LVCSR tasks to understand best practices and important design decisions when
building DNN acoustic models.

Recent research on DNN acoustic models for LVCSR explores variations
in network architecture, optimization techniques, and acoustic model
training loss functions. Due to system differences across research
groups it can be difficult, for example, to determine whether a
performance improvement is due to a better neural network architecture
or a different optimization technique. Our work aims to address these
concerns by systematically exploring several strategies to improve DNN
acoustic models. We view the acoustic modeling DNN component as a DNN
classifier and draw inspiration from recent DNN classification
research on other tasks -- predominantly image classification. Unlike
many other tasks, DNN acoustic models in LVCSR are not simply
classifiers, but are instead one sub-component of the larger speech
transcription system. There is a complex relationship between
downstream task performance, word error rate (WER), and the proximal
task of training a DNN acoustic model as a classifier. Because of this
complexity, it is unclear which improvements to DNN acoustic models
will ultimately result in improved performance across a range of LVCSR
tasks. 

This work empirically examines several aspects of DNN acoustic models
in an attempt to establish a set of best practices for creating such
models. Further, we seek to understand which aspects of DNN training
have the most impact on downstream task performance. This knowledge can
guide rapid development of DNN acoustic models for new speech corpora,
languages, computational constraints, and language understanding task
variants. Furthermore, we not only analyze task performance, but also
quantify differences in how various DNNs transform and represent
data. Understanding how DNNs process information helps us understand
underlying principles to further improve DNNs as classifiers and
components of large artificial intelligence systems. To this end, our
work serves as a case study for DNNs more generally as both
classifiers and components of larger systems.

We first perform DNN experiments on the standard Switchboard
corpus. We use this corpus to analyze the effect of DNN size on task
performance, and find that although there are 300 hours of training
data we can cause DNNs to over-fit on this task by increasing DNN
model size. We then investigate several techniques to reduce
over-fitting including the popular dropout regularization
technique. We next analyze neural network architecture choices by
comparing deep convolutional neural networks (DCNNs), deep locally
untied neural networks (DLUNNs), and standard DNNs. This comparison
also evaluates alternative input features since convolutional
approaches rely on input features with meaningful time and frequency
dimensions.

To explore DNN performance with fewer constraints imposed by
over-fitting, we next build a baseline LVCSR system by combining the
Switchboard and Fisher corpora. This results in roughly 2,100 hours of
training data and represents one of the largest collections of
conversational speech available for academic research. This larger
corpus allows us to explore performance of much larger DNN models, up
to ten times larger than those typically used for LVCSR. Using this
larger corpus we also evaluate the impact of optimization algorithm
choice, and the number of hidden layers used in a DNN with a fixed
number of total free parameters. We analyze our results not only in
terms of final task performance, but also compare sub-components of
task performance across models. Finally, we quantify differences in
how different DNN architectures process information.

Section~\ref{sec:nn_models} outlines the steps involved in building
neural network acoustic models for LVCSR, and describes previous work
on each step. This process outline contextualizes the questions
addressed by our investigations, which we present in
Section~\ref{sec:questions}. Section~\ref{sec:computations} describes
the neural network architectures and optimization algorithms evaluated
in this paper. Section~\ref{sec:experiment_swbd} presents our
experiments on the Switchboard corpus, which focus on regularization
and network dense versus convolutional architectural choices. We then
present experiments on the combined Switchboard and Fisher corpora in
Section~\ref{sec:experiment_combined} which explore the performance of
larger and deeper DNN architectures. We compare and quantify DNN
representational properties in Section~\ref{sec:coding_properties},
and conclude in Section~\ref{sec:conclusion}.

%% file: motivation.tex
\section{Neural Network Acoustic Models}
\label{sec:nn_models}
DNNs act as acoustic models for hidden Markov model (HMM) speech
recognition systems using the \emph{hybrid HMM} approach. A hybrid HMM
system largely resembles the standard HMM approach to speech
recognition using Gaussian mixture model (GMM) acoustic models.  A
full overview of LVCSR systems is beyond the scope of this work, so we
instead refer to previous articles for an overview of HMM-based speech
recognition systems
\cite{Gales2008,Young1997,Saon2012,Gold2011,Jurafsky2000}.

Our work focuses on the acoustic modeling component of the LVCSR system. 
The acoustic model approximates the distribution $p(x | y)$ which is
the probability of observing a given short span of acoustic features,
$x$, conditioned on an HMM state label, $y$. The acoustic input
features represent about 25ms of audio in most LVCSR systems. The HMM
state labels $y$ for LVCSR are \emph{senones} -- clustered,
context-dependent sub-phonetic states. A hybrid HMM system uses a
neural network to approximate $p(x | y)$ in place of a GMM. 

A neural network does not explicitly model the distribution $p(x | y)$
required by the HMM. Instead, we train neural networks to estimate
$p(y | x)$, which allows us to view the neural network as a classifier
of senones given acoustic input. We can use Bayes' rule to obtain $p(x
| y)$ given the neural network output distribution $p(y | x)$,
\begin{align}
  p(x | y ) &= \frac{p( y | x ) p(x)}{p(y)}.
\label{eqn:bayes}
\end{align}
The distribution $p(y)$ is the prior distribution over senones, which
we approximate as the empirical distribution of senone occurrence in
the training set. This is easy to obtain as it is simply a normalized
count of senones in the training set. We usually can not tractably
estimate probability of acoustic features, $p(x)$. This represents the
probability of observing a particular span of acoustic features -- a
difficult distribution to model. However, because our acoustic
features $x$ are fixed during decoding the term $p(x)$ is a constant,
albeit unknown, scaling factor. As a result we drop the term and
instead provide the HMM with an unscaled acoustic model score,
\begin{align}
  \frac{p( y | x )}{p(y)}.
\label{eqn:acoustic_score}
\end{align}
This term is not a properly formed acoustic model probability, but it
is sufficient to perform HMM decoding to maximize a combination of
acoustic and language model scores. The decoding procedure 
introduces an acoustic model scaling term to empirically adjust for
the scaling offset introduced by using un-normalized
probabilities.

Using neural networks as acoustic models for HMM-based speech
recognition was introduced over 20 years ago
\cite{Bourlard93,Renals1994,McClelland1986}. Much of this original
work developed the basic ideas of hybrid HMM-DNN systems which are
used in modern, state-of-the-art systems. However, until much more
recently neural networks were not a standard component in the highest
performing LVCSR systems. Computational constraints and the amount of
available training data severely limited the pace at which it was
possible to make progress on neural network research for speech
recognition. Instead, Gaussian mixture models were the standard
choice for acoustic modeling as researchers worked to refine the HMM
architecture, decoding frameworks, and signal processing challenges
associated with building high-performance speech recognizers.

While GMMs and their extensions produced gains on benchmark LVCSR
tasks over the span of many years, the resulting systems became
increasingly complex. Many of the complexities introduced focused
purely on increasing the representational capacity of GMM acoustic
models. In parallel to this effort, there was a resurgence of interest
in neural networks under the new branding of \emph{deep learning}
within the machine learning community. Work in this area focused on
overcoming optimization issues involved in training DNNs by applying
unsupervised pre-training to obtain a better initialization for
supervised learning tasks \cite{Hinton2006,Vincent2010}.

DNNs provided an interesting path forward for acoustic modeling as
neural networks offer a direct path to increasing representational
capacity, provided it is possible to find a good set of DNN
parameters. Early experiments with DNNs used fairly small phoneme
recognition tasks using monophone recognition systems and small
datasets like TIMIT~\cite{Mohamed2010}. In 2011 researchers
demonstrated that DNNs can also be applied to LVCSR systems with
context-dependent triphone states, rather than monophone states. This
innovation, coupled with the larger representational capacity of DNNs
as compared to GMMs, yielded substantial reductions in WER on multiple
challenging LVCSR tasks \cite{Dahl2011,Jaitly2012}. Within two years
DNN acoustic models showed gains on challenging tasks within the LVCSR
systems of Microsoft, Google, and IBM \cite{Hinton2012}.

Several factors are attributed to the success of modern DNN approaches
as compared to previous work with hybrid acoustic models.
Specifically the large total number of network parameters, increased
number of hidden layers, and initialization by pre-training were
thought to drive performance of modern hybrid HMM systems.
Researchers quickly established that hybrid HMMs work much better when
using context-dependent triphones in place of monophones
\cite{Dahl2011a}. Initializing DNN weights with unsupervised
pre-training was initially thought to be important for good
performance, but researchers later found that purely supervised
training from random initial weights yields nearly identical final
system performance~\cite{Yu2015}.  Using DNNs with many hidden layers
and many total parameters has generally found to be beneficial
\cite{Yu2013}, but the role of hidden layers and total network size is
not generally understood.

Having defined our hybrid HMM system and how we use the neural network
output $p(y|x)$ within the complete LVCSR system, we next focus on how
we build neural networks to model the senone distribution $p(y|x)$.
To better understand the detailed aspects related to building and
using neural network acoustic models for LVCSR we break the process
into a series of modeling and algorithmic choices. This set of steps
allows us to better contextualize previous work, and further convey
what aspects of the process are not yet fully understood. We define
the process as five steps:
\begin{enumerate}
\item
  {\bf Label Set}. The set of labels for our acoustic model are
  defined by the baseline HMM-GMM system we choose to use. Early work
  in neural network acoustic models used context-independent monophone
  states. Recent work with DNN acoustic models established that
  context-dependent states are critical to success~\cite{Dahl2011},
  which is generally true of modern LVCSR systems. Several variants of
  context-dependent states exist, and have been tried with DNN
  acoustic models. In this work we use context-dependent triphone
  senones created by our baseline HMM-GMM system.

\item
  {\bf Forced Alignment}. Our training data originally contains
  word-level transcriptions without time alignments for words.  We
  must assign a senone label to each acoustic input frame in each
  training utterance. We use a forced alignment of the ground-truth
  transcriptions to generate a sequence of senone labels for each
  utterance which is consistent with the word transcription for the
  utterance. Generating a forced alignment is a standard step of
  training any HMM-based system. The standard approach to hybrid
  speech recognition creates a forced alignment of the training data
  using an HMM-GMM system \cite{Gold2011}. The aligned data is then
  used to train a neural network acoustic model. Previous work found
  that using a trained HMM-DNN system to realign the training data for
  a second round of DNN training produces small gains in overall
  system performance \cite{Seide2011}. This process has more recently
  been generalized to yield an HMM-DNN training procedure which starts
  with no forced alignment but repeatedly uses a DNN to realign the
  training data \cite{Senior2014}. In our experiments we used a single
  forced alignment produced by the baseline HMM-GMM system as this the
  most standard approach when building DNN acoustic models.

\item
  {\bf Neural Network Architecture}. The size and structure of neural
  networks used for acoustic modeling is by far the largest difference
  between modern HMM-DNN systems and those used before 2010. Modern
  DNNs use more than one hidden layer, making them \emph{deep}. As a
  general property, depth is an important feature for the success of
  modern DNNs. Several groups recently found replacing the standard
  sigmoidal hidden units with rectified linear units in DNNs leads to
  WER gains and simpler training procedures for deep architectures
  \cite{Dahl2013,Zeiler2013,Maas2013}.

  Neural networks with only a single hidden layer perform
  worse than their deeper counterparts on a variety of speech tasks, even
  when the total number of model parameters is held
  fixed~\cite{Seide2011,Yu2013,Morgan2012}. Whether deeper is always better, or
  how deep a network must be to obtain good performance, is not well
  understood both for speech recognition and DNN classification tasks
  more generally. The total number of parameters used in modern DNNs
  is typically 10 to 100 times greater than neural networks used in
  the original hybrid HMM experiments. This increased model size,
  which translates to increased representational capacity, is critical
  to the success of modern DNN-HMM system. It is not clear how far we
  can push DNN model size or depth to continue increasing LVCSR
  performance.

  Size and depth are the most fundamental architectural choices for
  DNNs, but we can also consider a variety of alternative neural
  network architectures aside from a series of densely-connected
  hidden layers. DCNNs are an alternative to densely-connected
  networks which are intended to leverage the meaningful time and
  frequency dimensions in certain types of audio input
  features. Recent work with DCNNs found them to be useful first on
  phoneme recognition tasks but also on LVCSR tasks when used in
  addition to a standard DNN acoustic model
  \cite{AbdelHamid2012,Sainath2013,Sainath2014}. DCNNs change the
  first and sometimes second hidden layers of the neural network
  architecture, but otherwise utilize the same densely-connected
  multilayer architecture of DNNs. 

  Perhaps a larger architectural change from DNNs are deep
  \emph{recurrent} neural networks (DRNNs) which introduce a
  temporally recurrent hidden layer between hidden layers. The
  resulting architecture has outputs which no longer process each
  input context window independently, reflecting the temporal
  coherence and correlation of speech signals. DRNNs are a modern
  extension of the time-delay neural network first used for phoneme
  recognition by~\citet{Waibel1989} and recurrent network approach
  of~\citet{Robinson1991}. Modern recurrent network approaches to
  acoustic modeling have shown some initial success on large
  vocabulary tasks \cite{Sak2014}, and tasks where limited training
  data is available
  \cite{Sak2014a,Graves2013a,Vinyals2012,Weng2014}. The long term
  impact of DRNNs for HMM-DRNN systems is not yet clear as both the
  DRNN and HMM reason about the temporal dynamics of the input, which
  may introduce redundancy or interference. Researchers continue to
  propose and compare many architectural variants for acoustic
  modeling and other speech-related tasks \cite{Deng2013}.

\item
  {\bf Neural Network Loss Function}. Given a training set of
  utterances accompanied by frame-level senone labels we must choose a
  loss function to use when training our acoustic model. The space of
  possible loss functions is large, as it also includes the set of
  possible regularization terms we might use to control over-fitting
  during training. The default choice for DNN acoustic models is the
  cross entropy loss function, which corresponds to maximizing the
  likelihood of the observed label given the input. Cross entropy is
  the standard choice when training DNNs for classification tasks, but
  it ignores the DNN as a component of the larger ASR system. To
  account for more aspects of the overall system, discriminative loss
  functions were introduced for ASR tasks. Discriminative loss
  functions were initially developed for GMM acoustic models
  \cite{Bahl1986,Povey2008,Valtchev1997,Kaiser2000}, but were recently
  applied to DNN acoustic model training
  \cite{Vesely2013,Kingsbury2012,Su2013}. Discriminative training of
  DNN acoustic models begins with standard cross entropy training to
  achieve a strong initial solution. The discriminative loss function
  is used either as a second step, or additively combined with the
  standard cross entropy function. We can view discriminative training
  as a task-specific loss function which produces a DNN acoustic model
  to better act as a sub-component of the overall ASR system.

  For whatever loss function we choose, we can additionally apply one
  or more regularization terms to form the final training objective
  function. Regularization is especially important for DNNs where we
  can easily increase models' representational capacity. The simplest
  form of regularization widely applied to DNNs is a weight norm
  penalty, most often used with an $\ell_2$-norm penalty. While
  generally effective, developing new regularization techniques for
  DNNs is an area of active research. Dropout regularization
  \cite{Hinton2012} was recently introduced as a more effective
  regularization technique for DNN training. Recent work applied
  dropout regularization for DNN acoustic models, and found it
  beneficial when combined with other architectural changes \cite{Dahl2013}.

\item
  {\bf Optimization Algorithm}. Any non-trivial neural network model
  leads to a non-convex optimization problem. Because of this, our
  choice of optimization algorithm impacts the quality of local
  minimum found during optimization. There is little we can say in the
  general case about DNN optimization since it is not possible to find
  a global minimum nor estimate how far a particular local minimum is
  from the best possible solution. The most standard approach to DNN
  optimization is stochastic gradient descent (SGD). There are many
  variants of SGD, and practitioners typically choose a particular
  variant empirically. While SGD provides a robust default choice for
  optimizing DNNs, researchers continue to work on improving
  optimization algorithms for DNNs. Nearly all DNN optimization
  algorithms in popular use are gradient-based, but recent work has
  shown that more advanced quasi-Newton methods can yield better
  results for DNN tasks generally \cite{Martens2010,Ngiam2011} as well
  as DNN acoustic modeling \cite{Kingsbury2012}. Quasi-Newton and
  similar methods tend to be more computationally expensive per update
  than SGD methods, but the improved optimization performance can
  sometimes be distributed across multiple processors more easily, or
  necessary for loss functions which are difficult to optimize well
  with SGD techniques. Recently algorithms like AdaGrad
  \cite{Duchi2011} and Nesterov's Accelerated Gradient (NAG) were
  applied to DNNs for tasks outside of speech recognition, and tend to
  provide superior optimization as compared to SGD while still being
  computationally inexpensive compared to traditional quasi-Newton
  methods \cite{Sutskever2013}. 

  Amount of time required for training is an important practical
  consideration for DNN optimization tasks. Several groups have
  designed and implemented neural network optimization procedures
  which utilize graphics processing units
  (GPUs) \cite{Oh2004,Luo2005,Raina2009}, clusters of dozens to
  hundreds of computers \cite{Dean2012,Chilimbi2014,Chung2014}, or
  clusters of GPUs \cite{Coates2013}. Indeed, training time of neural
  networks has been a persistent issue throughout history, researchers
  often utilized whatever specialized computing hardware was available
  at the time \cite{Ellis1999,Lindsey1995}. Modern parallelized
  optimization approaches often achieve a final solution of similar
  quality to a non-parallelized optimization algorithm, but are
  capable of doing so in less time, or for larger models, as compared
  to non-parallelized approaches.
\end{enumerate}

%% file: questions.tex
\section{Questions Addressed in This Work}
\label{sec:questions}
At each stage of neural network acoustic model design and training
there is a tremendous breadth and depth of prior work. Researchers
often focus on improving one particular component of this pipeline
while holding all other components fixed. Unfortunately, there is no
well-established baseline for the acoustic model building pipeline, so
performance improvements of, for example, a particular architectural
variant are difficult to assess from examining the literature. Our
examines the relative
importance of several acoustic model design and training decisions.
By systematically varying several critical design components we are
able to test the limits of certain architectural choices, and uncover
which variations among baseline systems are most relevant for LVCSR
performance. We specifically address the following questions in this
work:

\begin{enumerate}
\item
  What aspects of neural network architecture are most important for
  acoustic modeling tasks? We investigate total network size and
  number of hidden layers using two corpora to avoid overfitting as a
  confounding factor. We build DNNs with five to ten times the total
  number of free parameters of DNNs used in most previous work. We
  also compare optimization algorithms to test whether more modern
  approaches to stochastic gradient descent are a driving factor in
  building large DNN acoustic models.

  We additionally compare a much broader architectural choice --
  locally-connected models versus the standard densely-connected DNN
  models. Recent work has found improvements when using DCNNs combined
  with DNNs for acoustic modeling, or when applying DCNNs to audio
  features with sufficient pre-processing \cite{Sainath2014}. We use
  two types of input features to compare DNNs with DCNNs. We present
  the first experiments DLUNNs for acoustic modeling. DLUNNs are a
  generalized version of DCNNs which are still locally connected but
  learn different weights at each location in the input features
  rather than sharing weights at all locations.

\item
  How can we improve the test set generalization of DNN acoustic
  models? Our experiments on DNN architecture choices reveal that
  increasing model size easily leads to overfitting issues.
  We evaluate several modifications to DNN
  training to improve the generalization performance of large DNNs. We
  include dropout, a recently introduced regularization technique, as
  well as early stopping, which has been used in neural network
  training for many years. Finally, we propose and evaluate
  \emph{early realignment}, a training technique specific to acoustic
  modeling, as a path towards improving generalization performance.

\item
  Do large, deep DNNs differ from shallow, smaller DNNs in terms of
  phonetic confusions or information processing metrics?
  DNN acoustic models are clearly successful in
  application but we do not yet understand why they perform well, or
  how they might be improved. We analyze the WER and classification
  errors made by large DNN acoustic models to test what improvements
  in sub-tasks ultimately lead to overall system WER
  improvements. Further, we look at information encoding metrics to
  quantify how information encoding changes in larger or deeper DNNs.
\end{enumerate}

We address each of these questions in separate experiments using the
Switchboard 300 hour corpus and a combined 2,100 hour corpus when
appropriate for the experiment. In Section~\ref{sec:computations} we
describe the DNN, DCNN, and DLUNN architecture computations used in
this work. Section~\ref{sec:experiment_swbd} addresses questions of
model size and overfitting on the Switchboard corpus while
Section~\ref{sec:experiment_cnn} uses the same baseline Switchboard
system to compare DCNN and DLUNN architectures to DNNs and baseline
GMMs. Section~\ref{sec:experiment_combined} presents experiments using
the larger training corpus to explore issues of model size, DNN depth,
and optimization algorithm. Sections~\ref{sec:error_analysis}
and~\ref{sec:coding_properties} analyze the performance and coding
properties of DNNs trained on the large combined corpus to better
understand how large DNNs encode information, and integrate into LVCSR
systems.

%% file: model.tex
\section{Neural Network Computations}
\label{sec:computations}
To address the stated research questions we employ three different
classes of neural network architecture. Each architecture amounts to a
different set of equations to convert input features into a predicted
distribution over output classes. We describe here the specifics of
each architecture, along with the loss function and optimization
algorithms we use.

\subsection{Cross Entropy Loss Function}
All of our experiments utilize the cross entropy classification loss
function. For some experiments we apply regularization techniques in
addition to the cross entropy loss function to improve generalization
performance. Many loss functions specific to speech recognition tasks
exist, and are a topic of active research. We choose to focus only on
cross entropy because training with cross entropy is almost always the
first step, or an additional loss function criterion, when
experimenting with more task-specific loss functions.  Additionally,
the cross entropy loss function is a standard choice for
classification tasks, and using it allows our experiments to serve as
a case study for large scale DNN classification tasks more generally.

The cross entropy loss function does not consider each utterance in
its entirety. Instead it is defined over individual samples of
acoustic input $x$ and senone label $y$.  The cross entropy objective
function for a single training pair $(x,y)$ is,
\begin{align}
  - \sum_{k=1}^K 1 \{ y=k \} \log \hat{y}_{k},
\label{eqn:crossent}
\end{align}
where $K$ is the number of output classes, and $\hat{y}_{k}$ is the probability
that the model assigns to the input example taking on label $k$.

Cross entropy is a convex approximation to the ideal 0-1 loss for
classification. However, when training acoustic models perfect
classification at the level of short acoustic spans is not our
ultimate goal. Instead, we wish to minimize the word error rate (WER)
of the final LVCSR system. WER measures mistakes at the word level,
and it is possible to perfectly transcribe the words in an utterance
without perfectly classifying the HMM state present at each time
step. Constraints present in the HMM and word sequence probabilities
from the language model can correct minor errors in state-level HMM
observation estimates. Conversely, not all acoustic spans are of equal
importance in obtaining the correct word-level transcription. The
relationship between classification accuracy rate at the frame level
and overall system WER is complex and not well understood. In our
experiments we always report both frame-level error metrics and
system-level WER to elicit insights about the relationship between DNN
loss function performance and overall system performance.

\subsection{Deep Neural Network Computations}
\begin{figure}[tb]
    \centering
    \input{figure/dnn.tex}

\caption{A DNN with 5-dimensional input, 3-dimensional
hidden layers, and 7-dimensional output. Each hidden layer is fully connected
to the to the previous and subsequent layer.}

\label{fig:dnn}
\end{figure}
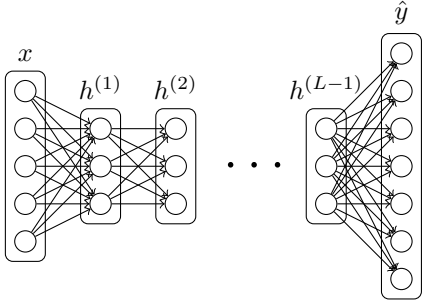

A DNN is a series of fully connected hidden layers which transform an
input vector $x$ into a probability distribution $\hat{y}$ to estimate
the output class. The DNN thus acts as a function approximator for the
conditional distribution $p(y | x)$. A DNN parametrizes this function
using $L$ layers, a series of hidden layers followed by an output
layer. Figure~\ref{fig:dnn} shows an example DNN. 

Each layer has a weight matrix $W$ and bias vector $b$. We
compute vector $h^{1}$ of first layer activations of a DNN using,
\begin{align}
  h^{(1)}(x) &= \sigma (W^{(1)T}x + b^{(1)}),
\end{align}
where $W^{(1)}$ and $b^{(1)}$ are the weight matrix and bias vectors
respectively for the first hidden layer. In this formulation each
column of the matrix $W^{(1)}$ corresponds to the weights for a single
hidden unit of the first hidden layer. Because the DNN is fully
connected, any real-valued matrix $W$ forms a valid weight matrix. If
we instead choose to impose partial connectivity, we are effectively
constraining certain entries in $W$ to be 0.

Subsequent hidden layers compute their hidden activation vector
$h^{(i)}$ using the hidden activations of the previous layer
$h^{(i-1)}$,
\begin{align}
  h^{(i)}(x) &= \sigma (W^{(i)T}h^{(i-1)} + b^{(i)}).
\end{align}

In all hidden layers we apply a point-wise nonlinearity function
$\sigma(z)$ as part of the hidden layer computation. Traditional
approaches to neural networks typically use a sigmoidal
function. However, in this work we use rectified linear units which
were recently shown to lead to better performance in hybrid speech
recognition as well as other DNN classification tasks\cite{Dahl2013,Zeiler2013,Maas2013}. 
The rectifier
nonlinearity is defined as,
\begin{align}
  \sigma(z) = \max (z, 0) = 
\begin{cases}
  z_i & z_i > 0 \\
  0 & z_i \leq 0
\end{cases}.
\label{eqn:unit_rel}
\end{align}

The final layer of the DNN must output a properly formed probability
distribution over the possible output categories. To do this, the
final layer of the DNN uses the \emph{softmax} nonlinearity, which is
defined as,
\begin{align}
  \hat{y}_j &= \frac{\exp(W^{(L)T}_j h^{(L-1)} + b^{(L)}_j)}
      {\sum_{k=1}^{N} \exp(W^{(L)T}_k h^{(L-1)} + b^{(L)}_k)}.
\end{align}

Using the softmax nonlinearity we obtain the output vector $\hat{y}$
which is a well-formed probability distribution over the $N$ output
classes. This distribution can then be used in the loss function
stated in Equation~\ref{eqn:crossent}, or other loss functions.

Having chosen a loss function and specified our DNN computation
equations, we can now compute a sub-gradient of the loss function with
respect to the network parameters. Note that because we are using
rectifier nonlinearities this is not a true
gradient, as the rectifier function is non-differentiable at 0. In
practice we treat this sub-gradient as we would a true gradient and
apply gradient-based optimization procedures to find settings for the
DNN's parameters.

This DNN formulation is fairly standard when compared to work in the
speech recognition community. The choice of rectifier nonlinearities
is a new one, but their benefit has been reproduced by several
research groups. Fully connected neural networks have been widely used
in acoustic modeling for over 20 years, but the issues of DNN total
size and depth have not been thoroughly studied.

\subsection{Deep Convolutional Neural Networks}
\label{sec:dcnn}

The fully-connected DNN architecture presented thus far serves as the
primary neural network acoustic modeling choice for modern speech
recognition tasks.  In contrast, neural networks for computer vision
tasks are often deep convolutional neural networks (DCNNs) which
exploit spatial relationships in input
data \cite{Lecun1998,Krizhevsky2012}.  When using spectrogram filter
bank representations of speech data, analogous time-frequency
relationships may exist. The DCNN architecture allows for parameter
sharing and exploiting local time-frequency relationships for improved
classification performance. DCNNs follow a convolutional layer with a
pooling layer to hard-code invariance to slight shifts in time and
frequency. Like fully connected neural network acoustic models, the
idea of using localized time-frequency regions for speech recognition
was introduced over 20 years ago \cite{Waibel1989}. Along with the
modern resurgence of interest in neural network acoustic models
researchers have taken a modern approach to DCNN acoustic models. Our
formulation is consistent with other recent work on DCNN acoustic
models \cite{Sainath2014}, but we do not evaluate specialized feature
post-processing or combining DNNs with DCNNs to form an ensemble of
acoustic models. Instead, we ask whether DCNNs should replace DNNs as
a robust baseline recipe for building neural network acoustic models.

\begin{figure*}[tb]

    \centering
    \includegraphics[width=0.9\textwidth]{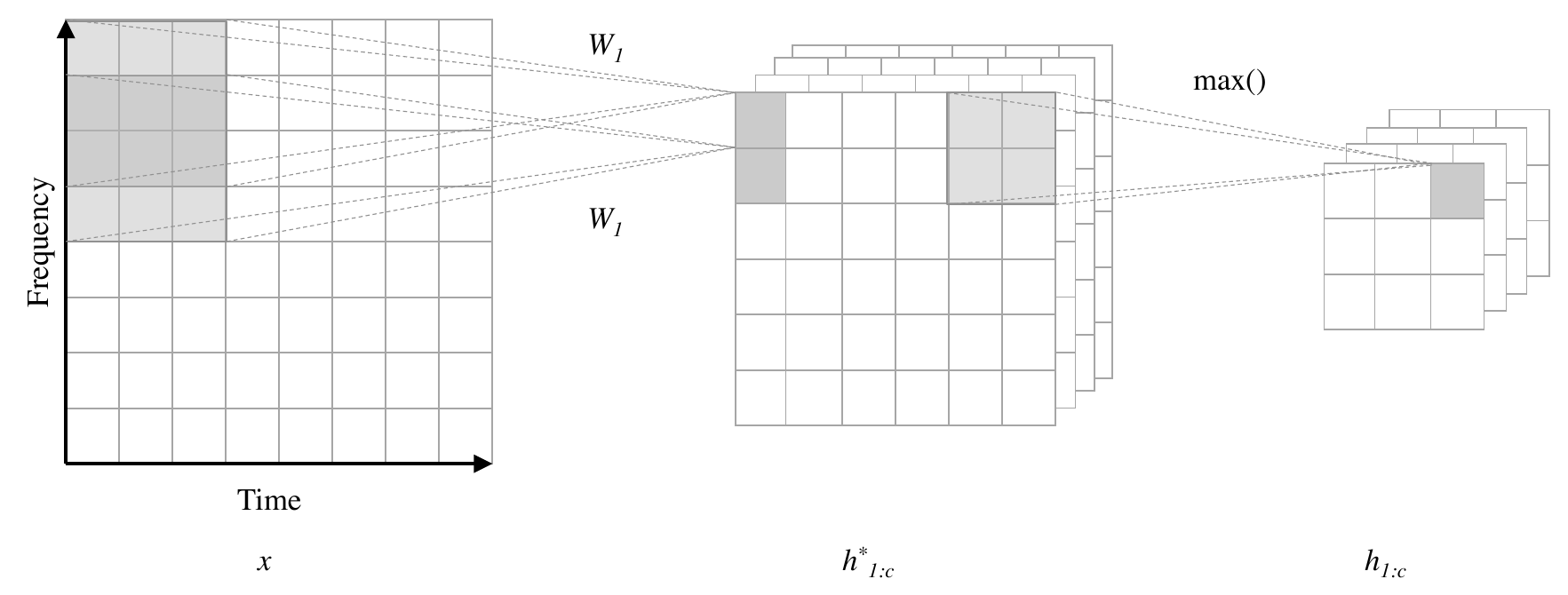}

\caption{Convolution and pooling first layer architecture. 
  Here the filter size is $5\times5$, and the pooling dimension
  is $3\times3$. Pooling regions are non-overlapping. Note that the
  $5\times5$ filters applied to each position in the convolution step
  are constrained to be the same. For max-pooling, the maximum value
  in each $3\times3$ grid is extracted.}

\label{fig:convnet}
\end{figure*}

Like a DNN, a DCNN is a feed-forward model which computes the
conditional distribution $p(y|x)$. The initial layers in a DCNN use
convolutional layers in place of the standard fully-connected layers
present in DNNs.  Convolutional layers were originally developed to
enable neural networks to deal with large image inputs for computer
vision tasks. In a convolutional model, we restrict the total number
of network parameters by using hidden units which connect to only a
small, localized region of the input. These localized hidden units are
applied at many different spatial locations to obtain hidden layer
representations for the entire input. In addition to controlling the
number of free parameters, reusing localized hidden units at different
locations leverages the stationary nature of many input domains. In
the computer vision domain, this amounts to reusing the same
edge-sensitive hidden units at each location of the image rather than
forcing the model to learn the same type of hidden unit for each
location separately.

Figure~\ref{fig:convnet} shows a convolutional hidden layer connected
to input features with time and frequency axes. A single weight matrix
$W_1$ connects to a 3x3 region of the input and we compute a
hidden unit activation value using the same rectifier nonlinearity
presented in Equation~\ref{eqn:unit_rel}. We apply this same procedure
at all possible locations of the input, moving one step at a time
across the input in both dimensions. This process produces a
\emph{feature map} $h^{(1,1)}$ which is the hidden activation values for
$W_1$ at each location of the input. The feature map itself has
meaningful time and frequency axes because we preserve these
dimensions as we convolve across the input to compute hidden unit
activations.

Our convolutional hidden layer has a feature map with redundancies
because we apply the hidden units at each location as we slide across
the input. Following the convolutional layer, we apply a
\emph{pooling} operation. Pooling acts as a down-sampling step, and
hard-codes invariance to slight translations in the input. Like the
localized windows used in the convolutional layer, the pooling layer
connects to a contiguous, localized region of its input -- the feature
map produced by a convolutional hidden layer. The pooling layer does
not have overlapping regions. We apply this pooling function to local
regions in each feature map. Recall that a feature map contains the
hidden unit activations for only a single hidden unit. We are thus
using pooling to select activation values for each hidden unit
separately, and not forcing different hidden units to compete with one
another. In our work, we use \emph{max pooling} which applies a max
function to the set of inputs in a single pooling region. Max pooling
is a common choice of pooling function for neural networks in both
computer vision and acoustic modeling tasks
\cite{Lee2009,AbdelHamid2012,Sainath2014}. The most widely used
alternative to max pooling replaces the max function with an averaging
function. Results with max pooling and average pooling are often
comparable.

The overall architecture of a DCNN consists of one or more layers of
convolution followed by pooling followed by densely connected hidden
layers and a softmax classifier. Essentially we build convolution and
pooling layers to act as input to a DNN rather than building a DNN
from the original input features. It is not possible to interleave
densely connected and convolutional hidden layers because a densely
connected hidden layer does not preserve spatial or time-frequency
relationships in their hidden layer representations.
The DCNN architecture contains more hyper-parameters than a standard
DNN because we must select the number of convolutional layers, input
region size for all convolution and pooling layers, and pooling
function. These are additional hyper-parameters to the choices of depth
and hidden layer size common to all types of deep neural network
architectures.

\subsection{Deep Local Untied Neural Networks}
\label{sec:dlunn}
\begin{figure*}[tb]
    \centering
    \includegraphics[width=0.9\textwidth]{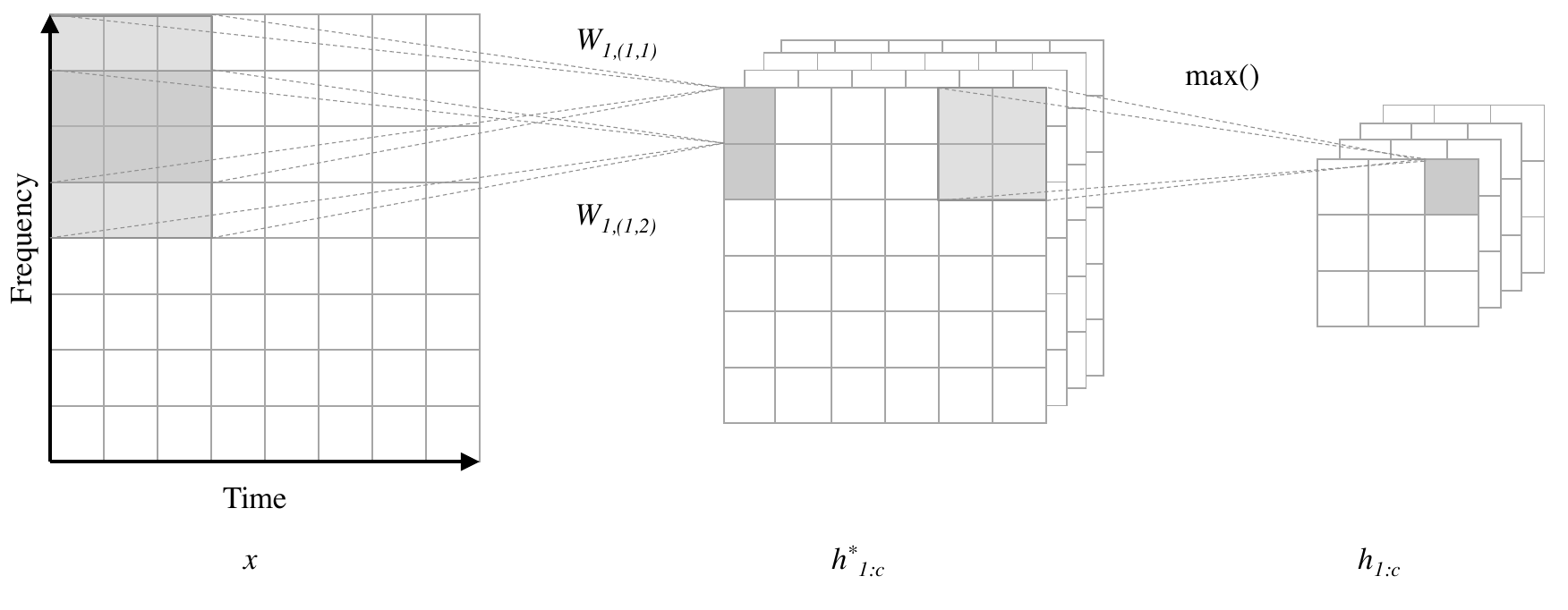}

\caption{Locally connected untied first layer architecture.  Here the
  filter size is $5\times5$, and the pooling dimension is
  $3\times3$. Pooling regions are non-overlapping. Unlike the
  convolutional layer shown in Figure~\ref{fig:convnet}, the network
  learns a unique $5\times5$ set of weights at each location. The max
  pooling layer otherwise behaves identically to the pooling layer in
  a convolutional architecture.}
\label{fig:dlunn}
\end{figure*}

DCNNs combine two architectural ideas simultaneously --
locally-connected hidden units and sharing weights across multiple
hidden units. We need not apply both of these architectural ideas
simultaneously. In a \emph{deep local untied neural network} (DLUNN)
we again utilize locally-connected hidden units but do not share
weights at different regions of the input. Figure~\ref{fig:dlunn}
shows an example DLUNN architecture, which differs only from a DCNN
architecture by using different weights at each location of the first
hidden layer. When applying a local untied hidden layer to
Mel-spectrum time-frequency input features the hidden units can
process different frequency ranges using different hidden units. This
allows the network to learn slight variations that may occur when a
feature occurs at a lower frequency versus a higher frequency.

In DLUNNs, the architecture is the same as in the convolutional
network, except that filters applied to different regions of the input
are not constrained to be the same. Thus untied neural networks can be
thought of as convolutional neural networks using locally connected
computations and without weight-sharing.  This results in a large
increase in the number of parameters for the untied layers relative to
DCNNs. Following each locally united layer we apply a max pooling
layer which behaves identically to the pooling layers in our DCNN
architecture. Grouping units together with a max pooling function
often results in hidden weights being similar such that the
post-pooling activations are an invariant feature which detects a
similar time-frequency pattern at different regions of the input. 
\subsection{Optimization Algorithms}
\label{sec:optimization}
Having defined several neural network architectures and the loss
function we wish to optimize, we must specify which gradient-based
algorithm we use to find a local minimum of our loss function. 
We consider only stochastic gradient techniques in our work as batch
optimization, which requires computing the gradient across the entire
dataset at each step, is impractical for the datasets we use.
There are several variants of stochastic gradient techniques, many with
different convergence properties when applied to convex optimization
problems. Because neural network training is a non-convex problem, it
is difficult to make general statements about optimality of
optimization methods. Instead, we consider the choice of optimization
algorithm as a heuristic which may lead to better performance in
practice. We consider two of the most popular stochastic gradient
techniques for our neural network training.

The first optimization algorithm we consider is stochastic gradient
with classical momentum (CM) \cite{Plaut1986,Rumelhart1985}. This technique is
probably the most standard optimization algorithm choice in modern
neural network research.  To minimize a cost function $f(\theta)$
classical momentum updates amount to,
\begin{align}
v_t &= \mu v_{t-1} - \epsilon \nabla f(\theta_{t-1})\\
\theta_t &= \theta_{t-1} + v_t,
\end{align}
where $v_t$ denotes the accumulated gradient update, or
\emph{velocity}, $\epsilon>0$ is the learning rate, and the momentum
constant $\mu\in[0,1]$ governs how we accumulate the velocity vector
over time. By setting $\mu$ close to one, one can expect to accumulate
the gradient information across a larger set of past updates. However,
it can be shown that for extremely ill-conditioned problems, a high
momentum for classical momentum method might actually cause
fluctuations in the parameter updates. This in turn can result in
slower convergence.

Recently the Nesterov's accelerated gradient (NAG) \cite{Nesterov1983}
technique was found to address some of the issues encountered when
training neural networks with CM. Both methods follow the intuition
that accumulating the gradient updates along the course of
optimization will help speed up convergence. NAG accumulates past
gradients using an alternative update equation that finds a better
objective function value with less sensitivity to optimization
algorithm hyper-parameters on some neural network tasks.
The NAG update rule is defined as, 
\begin{align}
v_t &= \mu_{t-1}v_{t-1} - \epsilon_{t-1}\nabla f(\theta_{t-1}+\mu_{t-1}v_{t-1})\\
\theta_t &= \theta_{t-1} + v_t.
\end{align}
Intuitively, this method avoids potential fluctuation in the
optimization by looking ahead to the gradient along the update
direction. For a more detailed explanation of the intuition underlying
NAG optimization for neural network tasks see Figure 7.1 in
\cite{Sutskever2013a}. In our work, we treat optimization algorithm
choice as an empirical question and compare CM with NAG on our
acoustic modeling task to establish performance differences.

%% file: figure/dnn.tex
\begin{tikzpicture}[transform shape]
    \tikzstyle{surround} = [draw=black,rounded corners=1mm]

    \def\numinp{5}
    \def\numhid{3}
    \def\numout{7}

    \foreach \k in {1,...,\numinp}{
        \node[draw,circle,inner sep=0.1cm, label=$ $] (i\k) at (0, 0.5*\k) {};
    }
    \node[surround, label=$x$] (ibbox) [fit = (i1) (i2) (i3) (i4) (i5)] {};

    \foreach \k in {1,...,\numhid}{
        \node[draw,circle,inner sep=0.1cm, label=$ $] (h1\k) at (1.0, 0.5+0.5*\k) {};
    }
    \node[surround, label=$h^{(1)}$] (h1bbox) [fit = (h11) (h12) (h13)] {};

    \foreach \k in {1,...,\numhid}{
        \node[draw,circle,inner sep=0.1cm, label=$ $] (h2\k) at (2.0, 0.5+0.5*\k) {};
    }
    \node[surround, label=$h^{(2)}$] (h2bbox) [fit = (h21) (h22) (h23)] {};

    \node[draw=none, label=$ $] (ellipsis1) at (3.1,1.5) {\scalebox{2}{$\cdots$}};

    \foreach \k in {1,...,\numhid}{
        \node[draw,circle,inner sep=0.1cm, label=$ $] (h3\k) at (4.0, 0.5+0.5*\k) {};
    }
    \node[surround, label=$h^{(L-1)}$] (h3bbox) [fit = (h31) (h32) (h33)] {};

    \foreach \k in {1,...,\numout}{
        \node[draw,circle,inner sep=0.1cm, label=$ $] (o\k) at (5.0, 0.5*\k-0.5) {};
    };
    \node[surround, label=$\hat{y}$] (h3bbox) [fit = (o1) (o2) (o3) (o4) (o5) (o6) (o7)] {};

    \foreach \ki in {1,...,\numinp}{
        \foreach \kh in {1,...,\numhid}{
            \path (i\ki) edge[->] (h1\kh);
        }
    }

    \foreach \jh in {1,...,\numhid}{
        \foreach \kh in {1,...,\numhid}{
            \path (h1\jh) edge[->] (h2\kh);
        }
    }

    \foreach \kh in {1,...,\numhid}{
        \foreach \ko in {1,...,\numout}{
            \path (h3\kh) edge[->] (o\ko);
        }
    }

\end{tikzpicture}

%% file: experiment_overfit.tex
\section{Switchboard 300 Hour Corpus}\label{sec:experiment_swbd}



We first carry out LVCSR experiments on the 300 hour Switchboard
conversational telephone speech corpus (LDC97S62). 
The baseline GMM system and forced alignments are created using the
Kaldi open-source toolkit\footnote{\tt{http://kaldi.sf.net}}
\cite{Povey2011}.  The baseline recognizer has 8,986 sub-phone states
and 200k Gaussians. The DNN is trained to estimate state likelihoods
which are then used in a standard hybrid HMM/DNN setup. Input features
for the DNNs are MFCCs with a context of $\pm 10$ frames.  Per-speaker
CMVN is applied and speaker adaptation is done using fMLLR. The
features are also globally normalized prior to training the DNN.
Overall, the baseline GMM system setup largely follows the existing
 {\tt s5b} Kaldi recipe and we defer to previous work for details
\cite{Vesely2013}.  For recognition evaluation, we report on a test set
consisting of both the Switchboard and CallHome subsets of the HUB5
2000 data (LDC2002S09) as well as a subset of the training set
consisting of 5,000 utterances. 

\subsection{Varying DNN Model Size}
\label{sec:exp_swbd_size}
We first experiment with perhaps the most direct approach to improving
performance with DNNs -- making DNNs larger by adding hidden
units. Increasing the number of parameters in a DNN directly increases
the representational capacity of the model. Indeed, this
representational scalability drives much of the modern interest in
applying DNNs to large datasets which might easily saturate other
types of models. Many existing experiments with DNN acoustic models
focus on introducing architecture or loss function variants to further
specialize DNNs for speech tasks. We instead ask the question of
whether model size alone can drive significant improvements in overall
system performance. We additionally experiment with using a larger
context window of frames as a DNN input as this should also serve as a
direct path to improving the frame classification performance of DNNs.

\subsubsection{Experiments}
We explore three different model sizes by varying the total number of
parameters in the network. The number of hidden layers is fixed to
five, so altering the total number of parameters affects the number of
hidden units in each layer. All hidden layers in a single network have
the same number of hidden units.  The hidden layer sizes are 2048,
3953 and 5984 which respectively yield models with approximately 36
million (M), 100M and 200M parameters. There are 8,986 output classes
which results in the output layer being the largest single layer in
any of our networks. In DNNs of the size typically studied in the
literature this output layer often consumes a majority of the total
parameters in the network. For example in our 36M parameter model the
output layer comprises 51\% of all parameters. In contrast, the output
layer in our 200M model is only 6\% of total parameters. Many output
classes occur rarely so devoting a large fraction of network
parameters to class-specific modeling may be wasteful. Previous work
explores factoring the output layer to increase the relative number of
shared parameters \cite{Senior2013,Sainath2013a}, but this effect
occurs naturally by substantially increasing network size. For our
larger models we experiment with the standard input of $\pm 10$ context
frames and additionally models trained with $\pm 20$ context frames.

All models use hidden units with the rectified linear nonlinearity.
For optimization, we use Nesterov's accelerated gradient with a smooth
initial momentum schedule which we clamp to a maximum of 0.95
\cite{Sutskever2013}. The stochastic updates are on mini-batches of 512
examples. After each epoch, or full pass through the data, we anneal
the learning rate by half. Training is stopped after improvement in
the cross entropy objective evaluated on held out development set
falls below a small tolerance threshold.


In order to efficiently train models of the size mentioned above, we
distribute the model and computation across several GPUs using the
distributed neural network infrastructure proposed by
\cite{Coates2013}. Our GPU cluster and distributed training software
is capable of training up to 10 billion parameter DNNs. We restrict
our attention to models in the 30M - 200M parameter range. In
preliminary experiments we found that DNNs with 200M parameters are
representative of DNNs with over one billion parameters for this task.
We train models for this paper in a model-parallel fashion by
distributing the parameters across four GPUs. A single pass through the
training set for a 200M parameter DNN takes approximately 1.5 days. 
Table \ref{tab:res_nonlin} shows frame-level and WER evaluations of
acoustic models of varying size compared against our baseline GMM
recognizer.

\subsubsection{Results}
Table~\ref{tab:res_nonlin} shows results for DNNs of varying size and
varying amounts of input context.  We find that substantially
increasing DNN size shows clear improvements in frame-level
metrics. Our 200M parameter DNN halves the development set cross
entropy cost of the smaller 36M parameter DNN -- a substantial
reduction. For each increase in DNN model size there is approximately
a 10\% absolute increase in frame classification accuracy. Frame-level
metrics are further improved by using larger context windows. In all
cases a model trained with larger context window outperforms its
smaller context counterpart. Our best overall model in terms of
frame-level metrics is a 200M parameter DNN with context window of
$\pm 20$ frames.

However, frame-level performance is not always a good proxy for WER
performance of a final system. We evaluate WER on a subset of the
training data as well as the final evaluation sets. Large DNN acoustic
models substantially reduce WER on the training set. Indeed, our
results suggest that further training set WER reductions are possible
by continuing to increase DNN model size. However, the gains we
observe on the training set in WER do not translate to large
performance gains on the evaluation sets. While there is a small
benefit of using models larger than the 36M DNN baseline size,
building models larger than 100M parameters does not prove beneficial
for this task.

\begin{table*}[bth]
\caption{Results for DNN systems in terms of frame-wise error metrics
  on the development set as well as word error rates on the training
  set and Hub5 2000 evaluation sets. The Hub5 set (EV) contains the
  Switchboard (SWBD) and CallHome (CH) evaluation subsets. We also
  include word error rates for the Fisher corpus development set (FSH)
  for cross-corpus comparison. Frame-wise error metrics were evaluated
  on 1.7M frames held out from the training set. DNN models differ
  only by their total number of parameters. All DNNs have 5 hidden
  layers with either 2,048 hidden units (36M parameters), 3,953 hidden
  units (100M parameters), or 5,984 hidden units (200M params).}
\label{tab:res_nonlin}
\begin{center}
\begin{small}
\begin{tabular}{lllrrrrrrr }
\toprule
Model Size & Layer Size & Context & Dev CrossEnt & Dev Acc(\%) & Train WER & SWBD WER & CH WER & EV WER\\
\midrule
GMM Baseline & N/A & $\pm 0$ & N/A & N/A & 24.93 & 21.7 & 36.1 & 29.0 \\
\midrule
36M & 2048 & $\pm 10$ & 1.23 & 66.20 & 17.52 & 15.1 & 27.1 & 21.2 \\
\midrule
100M & 3953 & $\pm 10$ & 0.77 & 78.56 & 13.66 & 14.5 & 27.0 & 20.8 \\
100M & 3953 & $\pm 20$ & 0.50 & 85.58 & 12.31 & 14.9 & 27.7 & 21.4 \\
\midrule
200M & 5984 & $\pm 10$ & 0.51 & 86.06 & 11.56 & 15.0 & 26.8 & 20.9 \\
200M & 5984 & $\pm 20$ & 0.26 & 93.05 & 10.09 & 15.4 & 28.5 & 22.0 \\
\bottomrule
\end{tabular}
\end{small}
\end{center}
\end{table*}

\subsubsection{Discussion}
\begin{figure}[tb]
\begin{center}
\centerline{\includegraphics[width=0.9\columnwidth]{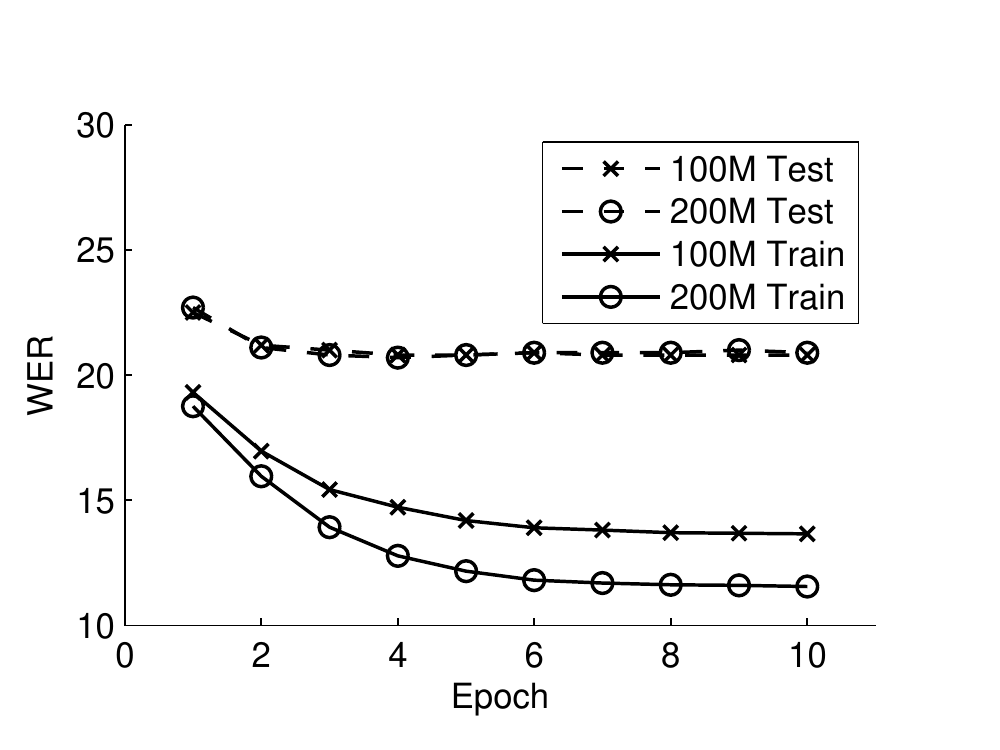}}
\caption{Train and test set WER as a function of training epoch for
  systems with DNN acoustic models of varying size. Each epoch is a
  single complete pass through the training set. Although the training
  error rate is substantially lower for large models, there is no gain
  in test set performance.}
\label{fig:overfit_wer}
\end{center}
\end{figure}

To better understand the dynamics of training large DNN acoustic
models, we plot training and evaluation WER performance during DNN
training. Figure~\ref{fig:overfit_wer} shows WER performance for our
100M and 200M parameter DNNs after each epoch of cross entropy
training. We find that training WER reduces fairly dramatically at
first and then continues to decrease at a slower but still meaningful
rate. In contrast, nearly all of our evaluation set performance is
realized within the first few epochs of training. This has two
important practical implications for large DNN training for speech
recognition. First, large acoustic models are not beneficial but do
not exhibit a strong over-fitting effect where evaluation set
performance improves for awhile before becoming increasingly
worse. Second, it may be possible to utilize large DNNs without
prohibitively long training times by utilizing our finding that most
performance comes from the first few epochs, even with models at our
scale. Finally, although increasing context window size improves all
training set metrics, those gains do not translate to improved test
set performance. It seems that increasing context window size provides
an easy path to better fitting the training function, but does not
result in the DNN learning a meaningful, generalizable function.

\subsection{Dropout Regularization}
Dropout is a recently-introduced technique to prevent over-fitting
during DNN training \cite{Hinton2012}. The dropout technique randomly
masks out hidden unit activations during training, which prevents
co-adaptation of hidden units. For each example observed during
training, each unit has its activation set to zero with probability $p
\in [0, 0.5]$.  Several experiments demonstrate dropout as a good
regularization technique for tasks in computer vision and natural
language processing \cite{Krizhevsky2012,Wager2013}. \cite{Dahl2013}
found a reduction in WER when using dropout on a 10M parameter DNN
acoustic model for a 50 hour broadcast news LVCSR task. Dropout
additionally yielded performance gains for convolutional neural
networks with less than 10M parameters on both 50 and 400 hour
broadcast news LVCSR tasks \cite{Sainath2013}. While networks which
employ dropout during training were found effective in these studies,
the authors did not perform control experiments to measure the impact
of dropout alone.  We directly compare a baseline DNN to a DNN of the
same architecture trained with dropout. This experiment tests whether
dropout regularization can mitigate the poor generalization
performance of large DNNs observed in Section~\ref{sec:exp_swbd_size}.


\subsubsection{Experiments}
We train DNN acoustic models with dropout to compare generalization
WER performance against that of the DNNs presented in
Section~\ref{sec:experiment_swbd}.  The probability of dropout $p$ is
a hyper-parameter of DNN training.  In preliminary experiments we
found setting $p=0.1$ to yield the best generalization performance
after evaluating several possible values, $p \in \{0.01, 0.1, 0.25,
0.5\}$. The DNNs presented with dropout training otherwise follow our
same training and evaluation protocol used thus far, and are built
using the same forced alignments from our baseline HMM-GMM system.

\subsubsection{Results}
Table~\ref{tab:wer_generalize} shows the test set performance of DNN
acoustic models of varying size trained with dropout.
DNNs trained with dropout improve over the baseline model for all acoustic
model sizes we evaluate. The improvement is a consistent 0.2\% to 0.4\%
reduction in absolute WER on the test set. While beneficial, dropout seems
insufficient to fully harness the representational capacity of our largest
models. Additionally, we note that hyper-parameter selection was critical to
finding any gain when using dropout. With a poor setting of the dropout
probability $p$ preliminary experiments found no gain and often worse results
from training with dropout. 
\begin{table}[bt]
\caption{Results for DNN systems trained with dropout regularization
  (DO) and early realignment (ER) to improve generalization
  performance. We build models with early realignment by starting
  realignment after each epoch starting after epoch two (ER2) and epoch
  five (ER5).  Word error rates are reported on the combined Hub5 test
  set (EV) which contains Switchboard (SWBD) and CallHome (CH)
  evaluation subsets. DNN model sizes are shown in terms of hidden
  layer size and millions of total parameters (e.g. 100M) }
\label{tab:wer_generalize}
\vskip 0.15in
\begin{center}
\begin{small}
\begin{tabular}{lrrr}
\toprule
Model & SWBD & CH & EV \\
\midrule
GMM Baseline & 21.7 & 36.1 & 29.0 \\
\midrule
2048 Layer (36M) & 15.1 & 27.1 & 21.2 \\
2048 Layer (36M) DO & 14.7 & 26.7 & 20.8 \\
\midrule
3953 Layer (100M) & 14.7 & 26.7 & 20.7 \\
3953 Layer (100M) DO & 14.6 & 26.3 & 20.5 \\
3953 Layer (100M) ER2 & 14.3 & 26.0 & 20.2 \\
3953 Layer (100M) ER5 & 14.5 & 26.4 & 20.5 \\
\midrule
5984 Layer (200M) & 15.0 & 26.9 & 21.0 \\
5984 Layer (200M) DO & 14.9 & 26.3 & 20.7 \\
\bottomrule
\end{tabular}
\end{small}
\end{center}
\vskip -0.1in
\end{table}
\subsection{Early Stopping}
Early stopping is a regularization technique for neural networks which
halts loss function optimization before completely converging to the
lowest possible function value. We evaluate early stopping as another
standard DNN regularization technique which may improve the
generalization performance of large DNN acoustic models.  Previous
work by \cite{Caruana2000} found that early stopping training of
networks with large capacity produces generalization performance on
par with or better than the generalization of a smaller
network. Further, this work found that, when using back-propagation for
optimization, early in training a large capacity network behaves
similarly to a smaller capacity network. Finally, early stopping as a
regularization technique is similar to an $\ell_2$ weight norm
penalty, another standard approach to regularization of neural network
training.

\subsubsection{Results}
By analyzing the training and test WER curves in
Figure~\ref{fig:overfit_wer} we can observe the best-case performance
of an early stopping approach to improving generalization. If we
select the lowest test set WER the system achieves during DNN
optimization, the 200M parameter DNN achieves 20.7\% WER on the EV subset -- 
only 0.1\% better than the
100M parameter baseline DNN system. This early stopped 200M model achieves only
a 0.5\% absolute WER reduction over the much smaller 36M parameter
DNN. This suggests that early stopping is beneficial, but perhaps
insufficient to yield the full possible benefits of large DNN acoustic
models.

\subsection{Early Realignment}
We next introduce a potential regularization technique which leverages
the process by which training labels are created for DNN acoustic
model training. Acoustic model training data is labeled via a forced
alignment of the word-level transcriptions. We test whether
re-labeling the training data \emph{during} training using the
partially-trained DNN leads to improved generalization performance.

Each short acoustic span $x_i$ has an associated HMM state label $y_i$
to form a supervised learning problem for DNN training. Recall that
the labels $y$ are generated by a forced alignment of the word-level
ground truth labels $w$ to the acoustic signal $x$. This forced
alignment uses an existing LVCSR system to generate a labeling $y$
consistent with the word-level transcription $w$. The system used to
generate the forced alignment is, of course, imperfect, as is the
overall speech recognition framework's ability to account for
variations in pronunciation. This leads to a dataset $\mathcal{D}$
where supervised training pairs $(x_i, y_i) \in \mathcal{D}$ contain
labels $y$ which are imperfect. We can consider a label $y_i$ as a
corrupted version of the true label $y_i^*$. The corruption function
which maps $y_i^*$ to $y_i$ is difficult to specify and certainly not
independent nor identically distributed at the level of individual
samples. Such a complex corruption function is difficult to analyze or
address with standard machine learning techniques for label noise. We
hypothesize, however, that the noisy labels $y$ are sufficiently
correct as to make significantly corrupted labels appear as outliers
with respect to the true labels $y^*$. Under this assumption we outline
an approach to improving generalization based on the 
dynamics of DNN performance during training optimization.

Neural networks exhibit interesting dynamics during optimization. Work
on early stopping found that networks with high capacity exhibit
behavior similar to smaller, limited capacity networks in early
phases of optimization \cite{Caruana2000}. Combining this finding with
the generally smooth functional form of DNN hidden and output units
suggests that early in training a large capacity DNN may fit a smooth
output function which ignores some of the label noise in $y$. Of
course, a large enough DNN should completely fit the corruptions
present in $y$ as optimization converges. Studies on the learning
dynamics of DNNs for hierarchical categorization tasks additionally
suggest that coarse, high-level output classes are fit first during
training optimization \cite{Saxe2013}.

Realignment, or generating a new forced alignment using an improved
acoustic model, is a standard tool for LVCSR system training.
Baseline LVCSR systems using GMM acoustic models
realign several times during training to iteratively improve.  While
iterative realignments have been helpful in improving system
performance in single-layer ANN-HMM hybrid models \cite{Bourlard93},
realignment is typically not used with large DNN acoustic models
because of the long training times of
DNNs. However, realignment using a fully trained DNN acoustic model often can
produce a small reduction in final system WER
\cite{Hinton2012}. 

We evaluate \emph{early realignment} which generates a new forced
alignment early in DNN optimization and then continues training on the
new set of labels.  Because large capacity DNNs begin accurately
predicting labels much earlier in training, early realignment may save
days of training time. Further, we hypothesize that a less fully
converged network can remove some label distortions while a more
completely trained DNN may already be fitting to the corrupt labels
given by an imperfect alignment.

\subsubsection{Experiments}
We begin by training an initial DNN using the same HMM-GMM forced
alignments and non-regularized training procedures presented thus far.
After training the DNN using the initial HMM-GMM alignments for a
fixed number of epochs, we use our new HMM-DNN system to generate a
new forced alignment for the entire training set. DNN training then
proceeds using the same DNN weights but the newly-generated training
set labels. As in our other regularization experiments, we hold the
rest of our DNN training and evaluation procedures fixed to directly
measure the impact of early realignment training. We train 100M
parameter five hidden layer DNNs and build models by realigning after
either two or five epochs. 

In preliminary experiments we found that realignment after each epoch
was too disruptive to DNN training and resulted in low quality DNN
models. Similarly, we found that starting from a fresh, randomly
initialized DNN after realignment performed worse than continuing
training from the DNN weights used to generate the realignment. We
found it important to reset the stochastic gradient learning rate to
its initial value after realignment occurs. Without doing so, our
annealing schedule sets the learning rate too low for the optimization
procedure to fully adjust to the newly-introduced labels. In a control
experiment, we found that resetting the learning rate alone, without
realignment, does not improve system performance.

\begin{figure}[tb]
\begin{center}
  \centerline{\includegraphics[width=0.9\columnwidth]{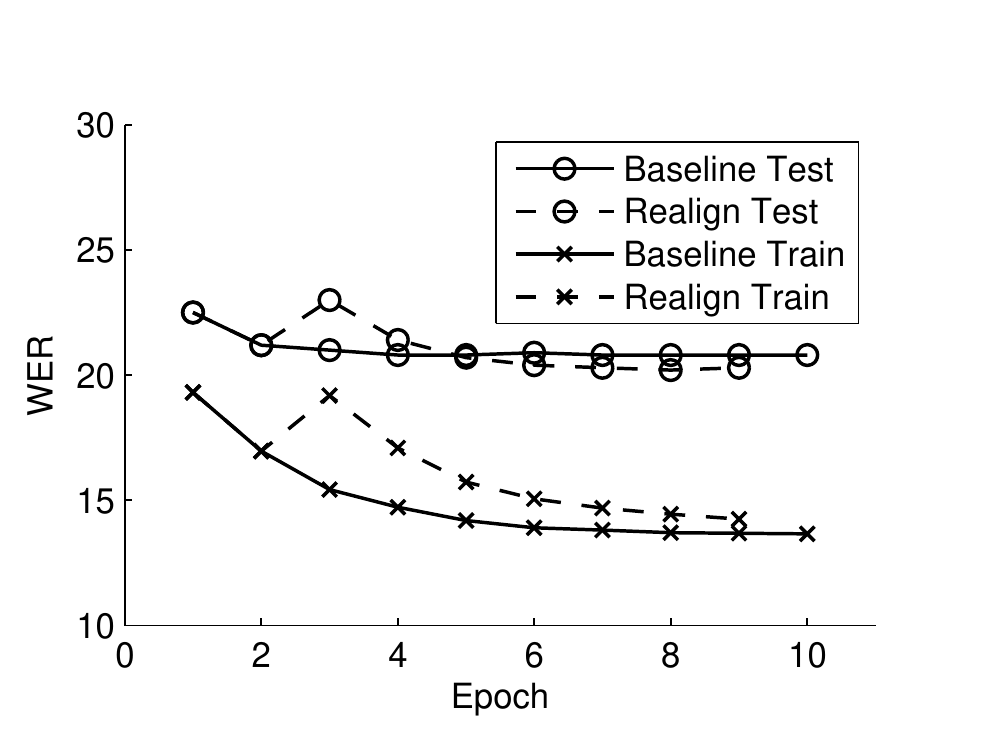}}
\caption{WER as a function of DNN training epoch for 
  systems with DNN acoustic models trained with and without label
  realignment after epoch 2. A DNN which re-generates its training
  labels with a forced alignment early during optimization generalizes
  much better to test data than a DNN which converges to the original
  labels.}
\label{fig:realign_wer}
\end{center}
\end{figure} 

\subsubsection{Results}
Table~\ref{tab:wer_generalize} compares the final test set performance
of DNNs trained with early realignment to a baseline model as well as
DNNs trained with dropout regularization. Realignment after five
epochs is beneficial compared to the baseline DNN system, but slightly
worse than a system which realigns after two epochs of training.
Early realignment leads to better WER performance than all models we
evaluated trained with dropout and early stopping. This makes early
realignment the overall best regularization technique we evaluated on
the Switchboard corpus. We note that only \emph{early} realignment
outperforms dropout regularization -- a DNN trained with realignment
after five epochs performs comparably to a DNN of the same size
trained with dropout.

\subsubsection{Discussion}
Figure~\ref{fig:realign_wer} shows training and test WER curves for
100M parameter DNN acoustic models trained with early realignment and
a baseline DNN with no realignment. We note that just after
realignment both train and test WER increase briefly. This is not
surprising as realignment substantially changes the distribution of
training examples. The DNN trained with realignment trains for three
epochs following realignment before it begins to outperform the
baseline DNN system.

We can quantify how much the labeling from realignment differs from
the original labeling by computing the fraction of labels changed. In
early realignment 16.4\% of labels are changed by realignment while
only 10\% of labels are changed when we realign with the DNN trained
for five epochs. This finding matches our intuition that as a large
capacity DNN trains it converges to fit the corrupted training samples
extremely well. Thus when we realign the training data with a fully
trained large capacity DNN the previously observed labels are
reproduced nearly perfectly. Realigning with a DNN earlier in
optimization mimics realigning with a higher bias model which relabels
the training set with a smoother approximate function. Taken
together, our results suggest early realignment leverages the high
bias characteristics of the initial phases of DNN training to reduce
WER while requiring minimal additional training time.

Early realignment also shows a huge benefit to training time compared
to traditional realignment. The DNN trained with realignment after
epoch five must train an additional three epochs, for a total of
eight, before it can match the performance of a DNN trained with early
realignment. For DNNs of the scale we use, this translates to several
days of compute time. The training time and WER reduction of DNNs with
early realignment comes with a cost of implementing and performing
realignment, which is of course not a standard DNN training
technique. Realignment requires specializing DNN training to the
speech recognition domain, but any modern LVCSR system should already
contain infrastructure to generate a forced alignment from an HMM-DNN
system. Overall, we conclude that early realignment is an effective
technique to improve performance of DNN acoustic models with minimal
additional training time.


%% file: experiment_cnn.tex
\section{Comparing DNNS, DCNNs, and DLUNNS on Switchboard}
\label{sec:experiment_cnn}


\begin{table*}[bt]
\caption{ Performance comparison of DNNs, deep convolutional neural
  networks (DCNNs), and deep local untied neural networks (DLUNNs).
  We evaluate convolutional models with one layer of convolution
  (DCNN) and two layers of convolution (DCNN2).  We compare models
  trained with fMLLR features and filter bank (FBank) features. Note
  that a context window of fMLLR features has a temporal dimension but
  no meaningful frequency dimension whereas FBank features have
  meaningful time-frequency axes. As an additional control we train a
  DCNN on features which are randomly permuted to remove meaningful
  coherence in both the time and frequency axes (FBank-P and
  fMLLR-P). We report performance on both the Hub5 Eval2000 test set
  (EV) which contains Switchboard (SWBD) and CallHome (CH) evaluation
  subsets. }
\label{tab:wer_cnn}
\begin{center}
\begin{small}
\begin{tabular}{llrrrr}
\toprule
Model & Features & Acc(\%) & SWBD WER & CH WER & EV WER \\
\midrule
GMM & fMLLR & N/A & 21.7 & 36.1 & 29.0 \\
\midrule
DNN & fMLLR  & 60.8 & 14.9 & 27.4 & 21.2 \\
DNN & FBank  & 51.7 & 16.5 & 31.6 & 24.1 \\
\midrule
DCNN & fMLLR  & 59.3 & 15.8 & 28.3 & 22.0 \\
DCNN & FBank  & 53.0 & 15.8 & 28.7 & 22.3 \\
\midrule
DCNN & fMLLR-P  & 59.0 & 15.9 & 28.6 & 22.4 \\
DCNN & FBank-P  & 50.7 & 17.2 & 32.1 & 24.7\\
\midrule
DCNN2 & fMLLR  & 58.8 & 15.9 & 28.3 & 22.2 \\
DCNN2 & FBank  & 53.0 & 15.6 & 28.3 & 22.1 \\
\midrule
DLUNN & fMLLR  & 61.2 & 15.2 & 27.4 & 21.3 \\
DLUNN & FBank  & 53.0 & 16.1 & 29.3 & 22.8 \\

\bottomrule
\end{tabular}
\end{small}
\end{center}
\end{table*}

The experiments thus far modify DNN training by adding various forms
of regularization. We now experiment with alternative neural network
architectures -- deep convolutional neural networks (DCNNs) and deep local
untied neural networks (DLUNNs).

\subsection{Experiments}
We trained DCNN and DLUNN acoustic models
using the same Switchboard training data as used for our DNN acoustic
model experiments to facilitate direct comparisons across
architectures. We evaluate filter bank features in addition to the fMLLR features
used in DNN training because filter bank features have meaningful
spectro-temporal dimensions for local receptive field
computations. All models have five hidden layers and were trained
using Nesterov's accelerated gradient with a smoothly increasing
momentum schedule capped at 0.95 and a step size of 0.01, halving the
step size after each epoch.

For our DCNN and DLUNN acoustic models we chose a receptive field of
$9\times9$ and non-overlapping pooling regions of dimension 
$1\times3$ (time by frequency).
Our models with two convolutional layers have the same first layer
filter and pooling sizes. The second layer uses a filter size of
$3\times3$  and does not use pooling. These parameters were selected using
results from preliminary experiments as well as results from previous
work \cite{Sainath2013}.

In the DCNNs one convolutional layer was used followed by four densely
connected layers with equal number of hidden units, and similarly for
the DLUNNs.  Map depth and number of hidden units were selected such
at all models have approximately 36M parameters. For DCNNs, the
convolutional first layer has a map depth of 128 applied to an input
with $\pm 10$ frame context. The following dense hidden layers each
have 1,240 hidden units. Our 2 convolutional layer DCNN uses 128
feature maps in both convolutional layers and 3 dense layers with
1,240 hidden units each. All DLUNNs use 108 filters at each location
in the first layer, and 4 hidden layers each with 1,240 hidden units.

The filter bank and
fMLLR features are both 40-dimensional. We ran initial experiments
convolving filters along frequency only, pooling along both frequency
and time, and overlapping pooling regions,
but did not find that these settings gave better performance. We ran
experiments with a context window of $\pm 20$ frames but found results
to be worse than results obtained with a context window of $\pm 10$
frames, so we report only the $\pm 10$ frame context results.

\subsection{Results}
Table~\ref{tab:wer_cnn} shows the frame-level and final system
performance results for acoustic models built from DNNs, DCNNs, and
DLUNNs.
When using filter bank features, DCNNs and DLUNNs both achieve
improvements over DNNs.  DCNN models narrowly outperform DLUNN
models. For locally connected acoustic models it appears that the
constraint of tied weights in convolutional models is advantageous as
compared to allowing a different set of localized receptive fields to
be learned at different time-frequency regions of the input. 

DLUNNs outperform DCNNs in experiments with fMLLR features. Indeed,
the DLUNN performs about as well as the DNN. The DCNN is harmed by the
lack of meaningful relationships along the frequency dimension of the
input features, whereas the more flexible architecture of the DLUNN is
able to learn useful first layer parameters. We also note that our
fMLLR features yield much better performance for all models as
compared with models trained on filter bank features.

In order to examine how much benefit using DCNNs to leverage local
correlations in the acoustic signal yields, we ran control experiments with
filter bank features randomly permuted along both the
frequency and time axes. The results show that while this harms
performance the convolutional architecture can still obtain fairly
competitive word error rates. This control experiment confirms that
locally connected models do indeed leverage localized properties of
the input features to achieve improved performance.

\subsection{Discussion}
While DCNN and DLUNN models are promising as compared to DNN models on
filter bank features, our results with filter bank features are
overall worse than results from models utilizing fMLLR features. 

Note that the filter bank features we used are fairly simple as
compared to our fMLLR features as the filter bank features do not
contain significant post-processing for speaker adaptation.  While
performing such feature transformations may give improved performance,
they call into question the initial motivation for using DCNNs to
automatically discover invariance to gender, speaker and
time-frequency distortions. The fMLLR features we compare against
include much higher amounts of specialized post-processing, which
appears beneficial for all neural network architectures we
evaluated. This confirms recent results from previous work, which
found that DCNNs alone are not typically superior to DNNs but can
complement a DNN acoustic model when both are used together, or
achieve competitive results when increased amounts of post-processing
are applied to filter bank features \cite{Sainath2014}. In summary, we
conclude that DCNNs and DLUNNs are not sufficient to replace DNNs as a
default, reliable choice for acoustic modeling network
architecture. We additionally conclude that DLUNNs warrant further
investigation as alternatives to DCNNs for acoustic modeling tasks.

%% file: experiment_combined_corpus.tex
\section{Combined Large Corpus}\label{sec:experiment_combined}
On the Switchboard 300 hour corpus we observed limited benefits from
increasing DNN model size for acoustic modeling, even with a variety
of techniques to improve generalization performance. We next explore
DNN performance using a substantially larger training corpus. This set
of experiments explores how we expect DNN acoustic models to behave
when training set size is not a limiting factor. In this setting,
over-fitting with large DNNs should be less of a problem and we can
more thoroughly explore architecture choices in large DNNs rather than
regularization techniques to reduce over-fitting and improve
generalization with a small training corpus.

\subsection{Baseline HMM system}\label{sec:baseline_combined}
To maximize the amount of training data for a conversational speech
transcription task, we combine the Switchboard corpus with the larger
Fisher corpus \cite{Cieri2004Fisher}. The Fisher corpus contains
approximately 2,000 hours of training data, but has transcriptions
which are slightly less accurate than those of the Switchboard corpus.

Our baseline GMM acoustic model was trained on features that are
obtained by splicing together 7 frames (3 on each side of the current
frame) of 13-dimensional MFCCs (C0-C12) and projecting down to 40
dimensions using linear discriminant analysis (LDA). The MFCCs are
normalized to have zero mean per speaker\footnote{This is done
  strictly for each individual speaker with our commit r4258 to the
  Kaldi recognizer. We found this to work slightly better than
  normalizing on a per conversation-side basis.}. After obtaining the
features with LDA, we also use a single semi-tied covariance (STC)
transform on the features. Moreover, speaker adaptive training (SAT)
is done using a single feature-space maximum likelihood linear
regression (fMLLR) transform estimated per speaker. The models trained
on the full combined Fisher+Switchboard training set contain 8725 tied
triphone states and 3.2M Gaussians.

The language model in our baseline system is trained on the
combination of the Fisher transcripts and the Switchboard Mississippi
State transcripts. Kneser-Ney smoothing was applied to fine-tune the
back-off probabilities to minimize the perplexity on a held out set of
10K transcript sentences from Fisher transcripts. In preliminary
experiments we interpolated the transcript-derived language model with
a language model built from a large collection of web page text, but
found no gains as compared with using the transcript-derived language
model alone.

We use two evaluation sets for all experiments on this corpus. First,
we use the same Hub5'00 (Eval2000) corpus used to evaluate systems on
the Switchboard 300hr task. This evaluation set serves as a reference
point to compare systems built on our combined corpus to those trained
on Switchboard alone. Second, we use the RT-03 evaluation set which is
more frequently used in the literature to evaluate Fisher-trained
systems. Performance of the baseline HMM-GMM system is shown in
Table~\ref{tab:wer_optimization} and Table~\ref{tab:wer_modelsize}.
\footnote{The implementation of our baseline HMM-GMM system is available in
the Kaldi project repository as example recipe \texttt{fisher\_swbd} (revision: r4340).}


\subsection{Optimization Algorithm Choice}\label{sec:experiment_optimization}
To avoid exhaustively searching over all DNN architecture and training
parameters simultaneously, we first establish the impact of
optimization algorithm choice while holding the DNN architecture
fixed.  We train networks with the two optimization algorithms
described in Section~\ref{sec:optimization} to determine which
optimization algorithm to use in the rest of the experiments on this
corpus.

\subsubsection{Experiments}
We train several DNNs with five hidden layers, where each layer has
2,048 hidden units. This results in DNNs with roughly 36M total free
parameters, which is a typical size for acoustic models used for
conversational speech transcription in the research literature.  For
both the classical momentum and Nesterov's accelerated gradient
optimization techniques the two key hyper-parameters are the initial
learning rate $\epsilon$ and the maximum momentum $\mu_{max}$. In all
cases we decrease the learning rate by a factor of 2 every 200,000
iterations. This learning rate annealing was chosen after preliminary
experiments, and overall performance does not appear to be
significantly affected by annealing schedule. It is more common to
anneal the learning rate after each pass through the dataset. Because
our dataset is quite large we found that annealing only after each
epoch leads to much slower convergence to a good optimization
solution.

\subsubsection{Results}
Table~\ref{tab:wer_optimization} shows both WER performance and
classification accuracy of DNN-based ASR systems with various
optimization algorithm settings. We first evaluate the effect of
optimization algorithm choice.  We evaluated DNNs with $\mu_{max} \in
{0.9, 0.95, 0.99}$ and $\epsilon \in \{0.1, 0.01, 0.001\}$.  For both
optimization algorithms DNNs achieve the best performance by setting
$\mu_{max}=0.99$ and $\epsilon = 0.01$.

In terms of frame level accuracy the NAG optimizer narrowly
outperforms the CM optimizer, but WER performance across all
evaluation sets are nearly identical. For both optimization algorithms
a high value of $\mu_{max}$ is important for good performance. Note
most previous work in hybrid acoustic models use CM with
$\mu_{max}=0.90$, which does not appear to be optimal in our
experiments. We also found that a larger initial learning rate was
beneficial. We ran experiments using $\epsilon \geq 0.05$ but do not
report results because the DNNs diverged during the optimization
process. Similarly, all models trained with $\epsilon = 0.001$ had WER
more than 1\% absolute higher on the EV test set as compared to the
same architecture trained with $\epsilon = 0.01$. We thus omit the
results for models trained with $\epsilon = 0.001$ from our results
table.

For the remainder of our experiments we use the NAG optimizer with
$\mu_{max}=0.99$ and $\epsilon=0.01$. These settings achieve the best
performance overall in our initial experiments, and generally we have
found the NAG optimizer to be somewhat more robust than the CM
optimizer in producing good parameter solutions.

\begin{table*}[bt]
\caption{ Results for DNNs of the same architecture trained with
  varying optimization algorithms. Primarily we compare stochastic
  gradient using classical momentum (CM) and Nesterov's accelerated
  gradient (NAG). We additionally evaluate multiple settings for the
  maximum momentum ($\mu_{max}$). The table contains results for only
  one learning rate ($\epsilon=0.01$) since it produces the best
  performance for all settings of optimization algorithm and
  momentum. We report performance on both the Hub5 Eval2000 test set
  (EV) which contains Switchboard (SWBD) and CallHome (CH) evaluation
  subsets. We also evaluate performance on the RT03 (RT03) Switchboard
  test set for comparison with Fisher corpus systems. }
\label{tab:wer_optimization}
\begin{center}
\begin{small}
\begin{tabular}{lrrrrrr}
\toprule
Optimizer & $\mu_{max}$ & Acc(\%) & SWBD WER & CH WER & EV WER & RT03 WER\\
\midrule
GMM & N/A & N/A & 21.9 & 31.9 & 26.9 & 39.5 \\
\midrule
CM & 0.90 & 52.51 & 18.3 & 27.3 & 22.8 & 39.0\\
CM & 0.95 & 54.20 & 17.1 & 25.6 & 21.4 & 38.1\\
CM & 0.99 & 55.26 & 16.3 & 24.8 & 20.6 & 37.5\\
\midrule
NAG & 0.90 & 53.18 & 18.0 & 26.7 & 22.3 & 38.5\\
NAG & 0.95 & 54.27 & 17.2 & 25.8 & 21.5 & 39.6\\
NAG & 0.99 & 55.39 & 16.3 & 24.7 & 20.6 & 37.4\\
\bottomrule
\end{tabular}
\end{small}
\end{center}
\end{table*}
\subsection{Scaling Total Number of DNN Parameters}
\label{sec:experiment_combined_scaling}
We next evaluate the performance of DNNs as a function of the total
number of model parameters while keeping network depth and
optimization parameters fixed. This approach directly assesses the
hypothesis of improving performance as a function of model size when
there is sufficient training data available. We train DNNs with 5
hidden layers, and keep the number of hidden units constant across
each hidden layer. Varying total free parameters thus corresponds to
adding hidden units to each hidden
layer. Table~\ref{tab:wer_modelsize} shows the frame classification
and WER performance of 5 hidden layer DNNs containing 36M, 100M, 200M,
and 400M total free parameters. Because it can be difficult to exactly
reproduce DNN optimization procedures, we make our DNN training code
available online~\footnote{For DNN training code, see \texttt{<upon
    acceptance>}}. Our DNN training code comprises only about 300
lines of Python code in total, which should facilitate easy comparison
to other DNN training frameworks.

Overall, the 400M parameter model performs best in terms of both frame
classification and WER across all evaluation sets. Unlike with our
smaller Switchboard training corpus experiments, increasing DNN model
size does not lead to significant over-fitting problems in
WER. However, the gain from increasing model size from 36M to 400M,
more than a 10x increase, is somewhat limited. On the Eval2000
evaluation set we observe a 3.8\% relative gain in WER from the 100M DNN as
compared to the 36M DNN. When moving from the 100M DNN to the 200M DNN
there is relative WER gain of 2.5\%. Finally the model size increase
from 200M to 400M total parameters yields a relative WER gain of
1\%. There are clearly diminishing returns as we
increase model size. The trend of diminishing relative gains in WER
also occurs on the RT03 evaluation set, although relative gains on
this evaluation set are somewhat smaller overall.

Frame classification rates on this corpus are much lower overall as
compared with our Switchboard corpus DNNs. We believe this corpus is
more challenging due to more overall acoustic variation, and errors
induced by quick transcriptions. Even our largest DNN leaves room for
improvement in terms of frame classification. In
Section~\ref{sec:error_analysis} we explore more thoroughly the
frame classification performance of the DNNs presented here.

\begin{table*}[bt]
\caption{ Results for DNNs of varying total model size and DNN depth.
  We report performance on both the Hub5 Eval2000 test set (EV) which
  contains Switchboard (SWBD) and CallHome (CH) evaluation subsets. We
  also evaluate performance on the RT03 (RT03) Switchboard test set
  for comparison with Fisher corpus systems. We additionally report
  frame-level classification accuracy (Acc) on a held out test set to
  compare DNNs as classifiers independent of the HMM decoder.}
\label{tab:wer_modelsize}
\begin{center}
\begin{small}
\begin{tabular}{rrr rrrrr}
\toprule
\# Params & Nun. Layers & Layer Size & Acc(\%) & SWBD WER & CH WER & EV WER & RT03 WER\\
\midrule
GMM & N/A & N/A & N/A & 21.9 & 31.9 & 26.9 & 39.5\\
\midrule
36M & 1 & 3803 & 49.38 & 21.0 & 30.4 & 25.8 & 43.2\\
36M & 3 & 2480 & 54.78 & 17.0 & 25.8 & 21.4 & 38.2\\
36M & 5 & 2048 & 55.37 & 16.2 & 24.7 & 20.6 & 37.4\\
36M & 7 & 1797 & 54.99 & 16.3 & 24.7 & 20.7 & 37.3\\
\midrule
100M & 1 & 10454 & 50.82 & 19.8 & 29.1 & 24.6 & 42.4\\
100M & 3 & 4940 & 56.02 & 16.3 & 24.8 & 20.6 & 37.3\\
100M & 5 & 3870 & 56.62 & 15.8 & 23.8 & 19.8 & 36.7\\
100M & 7 & 3309 & 56.59 & 15.7 & 23.8 & 19.8 & 36.4\\
\midrule
200M & 1 & 20907 & 51.29 & 19.6 & 28.7 & 24.3 & 42.8\\
200M & 3 & 7739 & 56.58 & 16.0 & 24.0 & 20.1 & 37.0\\
200M & 5 & 5893 & 57.36 & 15.3 & 23.1 & 19.3 & 36.0\\
200M & 7 & 4974 & 57.28 & 15.3 & 23.3 & 19.3 & 36.2\\
\midrule
400M & 5 & 8876 & 57.70 & 15.0 & 23.0 & 19.1 & 35.9\\
\bottomrule
\end{tabular}
\end{small}
\end{center}
\end{table*}

\subsection{Number of Hidden Layers}
\label{sec:experiment_depth}
We next compare performance of DNN systems while keeping total model
size fixed and varying the number of hidden layers in the DNN.
The optimal architecture for a neural network may change as the total
number of model parameters changes. There is no a priori reason to
believe that 5 hidden layers is optimal for all model
sizes. Furthermore, there are no good general heuristics to select the
number of hidden layers for a particular
task. Table~\ref{tab:wer_modelsize} shows DNN system performance for
DNNs with 1, 3, 5, and 7 hidden layers for DNNs of at multiple total
parameter counts.

The most striking distinction in terms of both frame classification
and WER is the performance gain of deep models versus
those with a single hidden layer.  Single
hidden layer models perform much worse than DNNs with 3 hidden layers
or more. Among deep models there are much smaller gains as a function
of depth. Models with 5 hidden layers show a clear gain over those
with 3 hidden layers, but there is little to no gain from a 7 hidden
layer model when compared with a 5 hidden layer model. These results
suggest that for this task 5 hidden layers may be deep enough to
achieve good performance, but that DNN depth taken further does not
increase performance. It's also interesting to note that DNN depth has
a much larger impact on performance than total DNN size. For this
task, it is much more important to select an appropriate number of
hidden layers than it is to choose an appropriate total model size.

For each total model size there is a slight decrease in frame
classification in 7 layer DNNs as compared with 5 hidden layer
DNNs. This trend of decreasing frame-level performance is also present
in the training set, which suggests that as networks
become very deep it is more difficult to minimize the training
objective function. This is evidence for a potential confounding
factor when building DNNs. In theory deeper DNNs should be able to
model more complex functions than their shallower counterparts, but in
practice we found that depth can act as a regularizer due to the difficulties in
optimizing very deep models.
\section{WER and Frame Classification Error Analysis}
\label{sec:error_analysis}
We now decompose our task performance metrics of frame classification
accuracy and WER into their constituent components to gain a deeper
understanding of how models compare to one another. This analysis
attempts to uncover differences in models which achieve similar
aggregate performance. For example, two systems which have the same
final WER may have different rates of substitutions, deletions, and
insertions -- the constituent components of the WER metric.

\begin{figure}[tb]
\begin{center}
\centerline{\includegraphics[width=1.0\columnwidth]{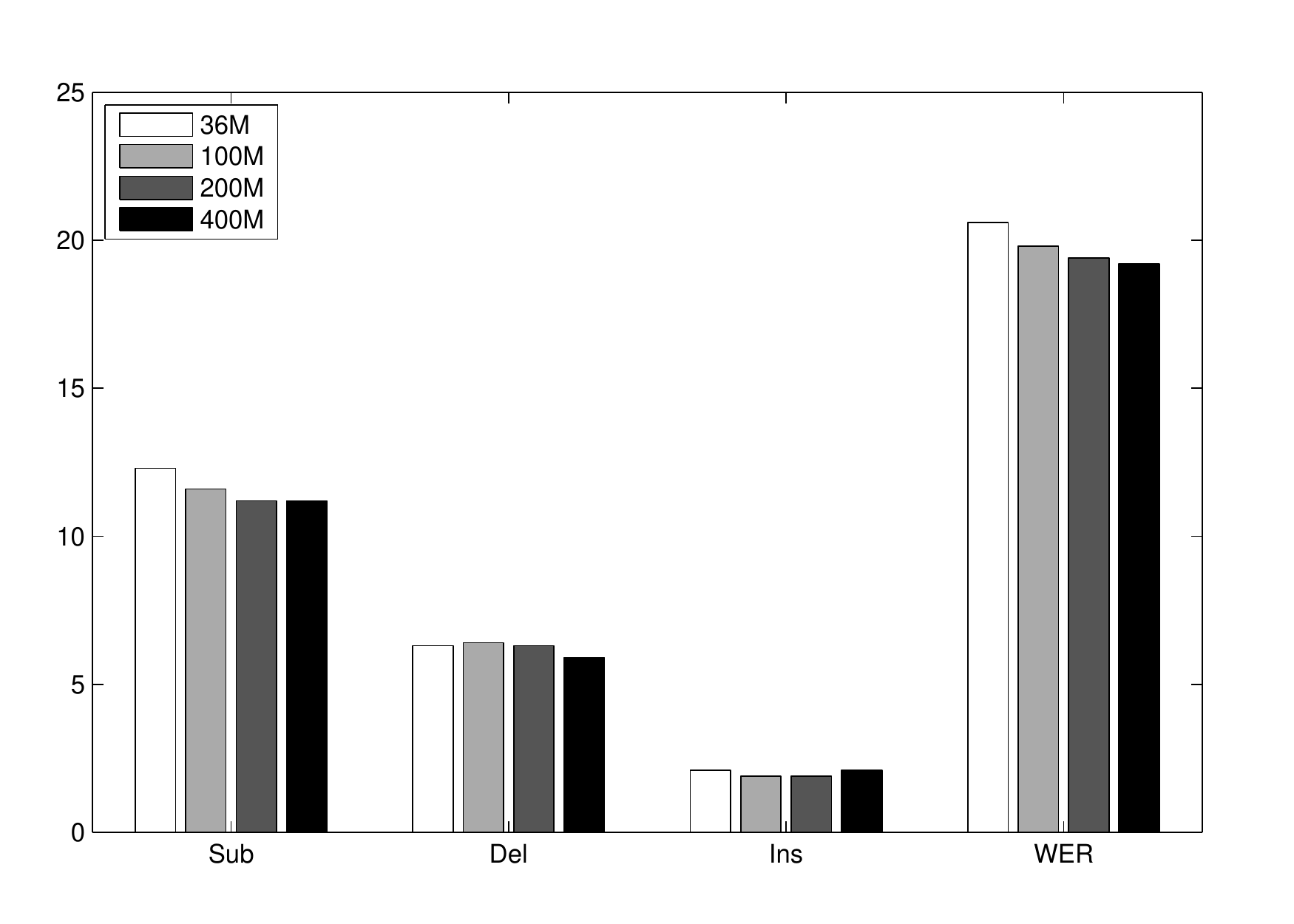}}
\caption{ Eval2000 WER of 5 hidden layer DNN systems of varying total parameter
  count. WER is broken into its sub-components -- insertions,
  substitutions, and deletions.}
\label{fig:wer_breakdown}
\end{center}
\end{figure}

Figure~\ref{fig:wer_breakdown} shows decomposed WER performance of
HMM-DNN systems of varying DNN size. Each HMM-DNN system uses a DNN
with 5 hidden layers, these are the same HMM-DNN systems reported in
Table~\ref{tab:wer_modelsize}. We see that decreases in overall WER as
a function of DNN model size are largely driven by lower substitution
rates. Insertions and deletions remain relatively constant across
systems, and are generally the smaller components of overall
WER. Decreased substitution rates should be a fairly direct result of
improving acoustic model quality as the system becomes more confident
in matching audio features to senones. While the three WER
sub-components are linked, it is possible that insertions and
deletions are more an artifact of other system shortcomings such as
out of vocabulary words (OOVs) or a pronunciation dictionary which
does not adequately capture pronunciation variations.

We next analyze performance in terms of frame-level classification
grouped by phoneme.  When understanding senone classification we can
think of the possible senone labels as leaves from a set of
trees. Each phoneme acts as the root of a different tree, and the
leaves of a tree correspond to the senones associated with a the
tree's base phoneme.

\begin{figure*}[tb]
\begin{center}
\centerline{\includegraphics[width=2.25\columnwidth]{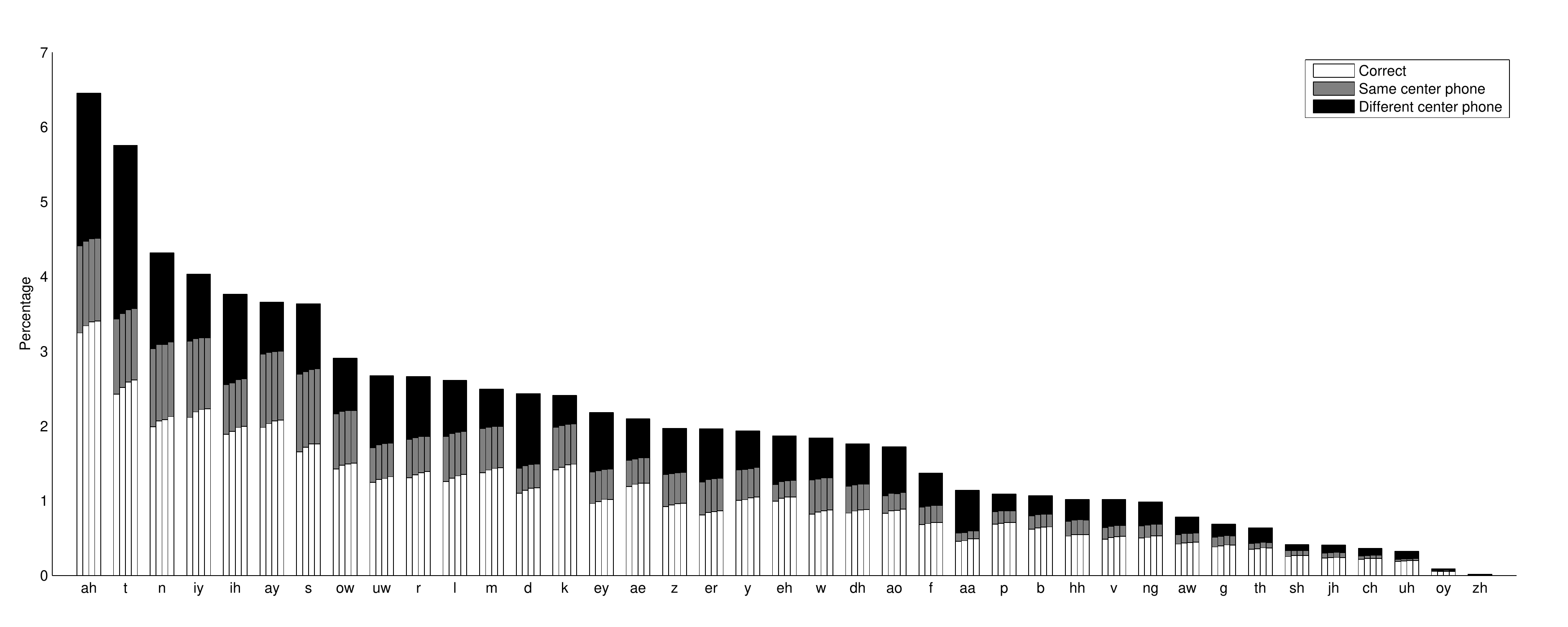}}
\caption{ Senone accuracy of 5 hidden layer DNN systems of varying
  total parameter count. Accuracy is grouped by base phone and we
  report the percentage correct, mis-classifications which chose a
  senone of the same base phone, and mis-classifications which chose a
  senone of a different base phone. The total size of the combined bar
  indicates the occurrence rate of the base phone in our data
  set. Each base phone has five bars, each representing the
  performance of a different five layer DNN. The bars show
  performance of DNNs of size 36M 100M 200M and 400M from left to
  right. We do not show the non-speech categories of silence,
  laughter, noise, or OOV which comprise over 20\% of frames sampled.}
\label{fig:phone_acc_breakdown}
\end{center}
\end{figure*}

Figure~\ref{fig:phone_acc_breakdown} shows classification percentages
of senones grouped by their base phoneme. The DNNs analyzed are the
same 5 hidden layer models presented in our WER analysis of
Figure~\ref{fig:wer_breakdown} and Table~\ref{tab:wer_modelsize}.  The
total height of each bar reflects its percentage of occurrence in our
data. Each bar is then broken into three components -- correct
classifications, errors within the same base phoneme, and errors
outside the base phoneme.  Errors within the base phoneme correspond
the examples where the true label is a senone from a particular base
phone, e.g. \emph{ah}, but the network predicts an incorrect senone
label also rooted in \emph{ah}. The other type of error possible is
predicting a senone from a different base phoneme. Together these
three categories, correct, same base phone, and different base phone,
additively combine to form the total set of senone examples for a
given base phone.

The rate of correct classifications is non-decreasing as a function of
DNN model size for each base phoneme. The overall increasing accuracy
of larger DNNs comes from small correctness increases spread across
many base phonemes. Across phonemes we see substantial differences in
within-base-phoneme versus out-of-base-phoneme error rates. For
example, the vowel \emph{iy} has a higher rate of within-base-phoneme
errors as compared to the fairly similar vowel \emph{ih}. Similarly,
the consonants \emph{m}, \emph{k}, and \emph{d} have varying rates of
within-base versus out-of-base errors despite having similar total
rates of base phoneme occurrence in the data. We note that our DNNs
generally exhibit similar error patterns to those observed with DNN
acoustic models on smaller corpora \cite{Huang2014}. However, due to
the challenging nature of our corpus we observe overall lower phone
accuracies than those found in previous work. Performance as a
function of model size appears to change gradually and fairly
uniformly across phonemes, rather than larger models improving upon
only specific phonemes, perhaps at the expense of performance on
others.

%% file: coding_properties.tex
\section{Analyzing Coding Properties}
\label{sec:coding_properties}
Our experiments so far focus on task performance at varying levels of
granularity. These metrics address the question of \emph{what} DNNs
are capable of doing as classifiers and when integrated with HMM
speech decoding infrastructure. However, we have not yet completely
addressed the question of \emph{how} various DNN architectures achieve
their various levels of task performance. While the DNN computation
equations presented in Section~\ref{sec:computations} describe the
algorithmic steps necessary to compute predictions, there are many
possible settings of the free parameters in a model. In this section
we offer a descriptive analysis of how our trained DNNs encode
information. This analysis aims to uncover quantifiable differences in
how models of various sizes and depths encode input data and transform
it to make a final prediction.

\subsection{Sparsity and Dispersion}
\begin{figure*}[tb]
\begin{center}
\centerline{\includegraphics[width=2.25\columnwidth]{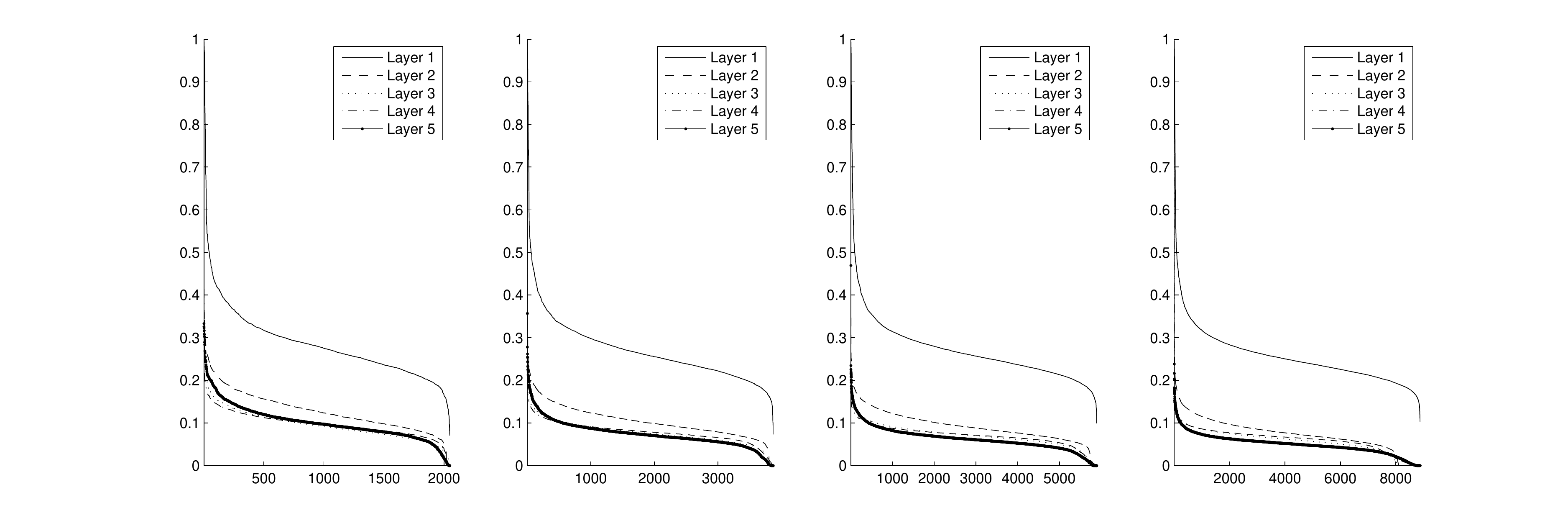}}
\caption{ Empirical activation probability of hidden units in each
  hidden layer layer of 5 hidden layer DNNs. Hidden units (x axis) are
  sorted by their probability of activation. We consider any positive
  value as active ($h(x)>0$). Each sub-figure corresponds to a
  different model size of 36M, 100M, and 200M total parameters from
  left to right. }
\label{fig:scree_5layer}
\end{center}
\end{figure*}

Our first analysis focuses on the sparsity patterns of units with each
hidden layer of a DNN. We compute the empirical \emph{lifetime
  sparsity} of each hidden unit by forward propagating a set of 512,000
examples through the DNN. We consider a unit as a active when its
output is non-zero, and compute the fraction of examples for which a
unit is active as its \emph{lifetime activation probability}. This
value gives the empirical probability that a particular unit will
activate given a random input drawn from our sample distribution. For
each hidden layer of a network, we can plot all hidden units' lifetime
activation probabilities sorted in decreasing order to get a sense for
the distribution of activation probabilities within a layer. This
plotting technique, sometimes called a scree plot, helps us understand
how information coding is distributed across units in a hidden
layer. Figure~\ref{fig:scree_5layer} shows a set of scree plots for 5
hidden layer DNNs of varying total model size.

From a coding theory perspective, researchers often discuss DNNs as
learning efficient codes which are both sparse and dispersed. Sparsity
generally refers to relatively few hidden units in a hidden layer
being active in response to an input. Sparsity is efficient and seems
natural given modern DNN structures in which hidden layer size is
often much larger than input vector dimensionality. Dispersion refers
to units within a hidden layer equally sharing responsibility for
coding inputs. A representation with perfect dispersion would appear
flat in a scree plot. A scree plot also visualizes sparsity as the
average height of representation units on the y axis.

Generally we see that in all model sizes sparsity increases in deeper
layers of the DNN. The first hidden layer is noticeably more active on
average as compared with every other layer in the DNN, in most cases
by almost a factor of two. Beyond the first layer, activation
probability per layer decreases slightly as we look at deeper layers
of the DNN. The changes in activation probability per layer within
deeper hidden layers are fairly minor, which suggest that a
representation is transformed but not continually compressed. 

Dispersion is similar within layers of a particular DNN
size. Generally the representations appear fairly disperse, with a
mostly flat curve for each hidden layer and only a few units which are
on or off for a large percentage of inputs at each tail. There does
appear to be a slight trend of increasing dispersion in deeper layers
of the DNN, especially in larger models. 

Most importantly, we do not observe a significant set of permanently
inactive units as DNNs grow in total number of parameters. In larger
DNNs the representation remains fairly disperse, with only a small set
of units which are active for less than 1\% of inputs. This is an
important metric because adding more parameters to a DNN is only
useful in so far as those parameters are actually used in encoding and
transforming inputs.

\begin{figure*}[tb]
\begin{center}
\centerline{\includegraphics[width=2.25\columnwidth]{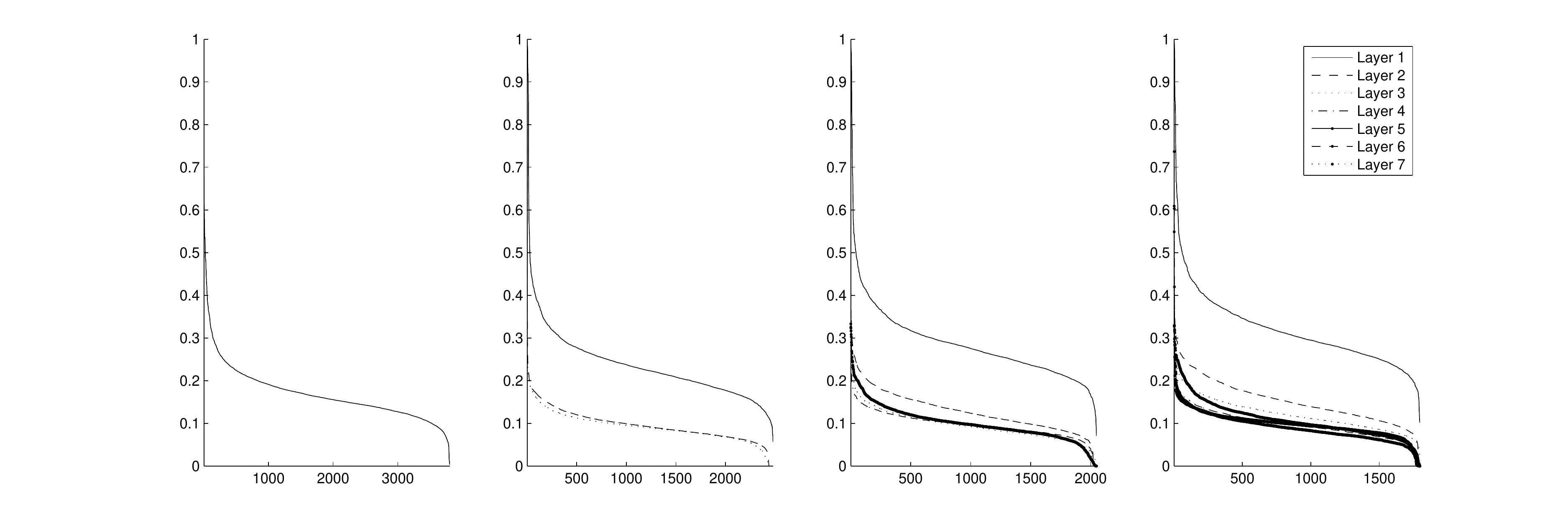}}
\centerline{\includegraphics[width=2.25\columnwidth]{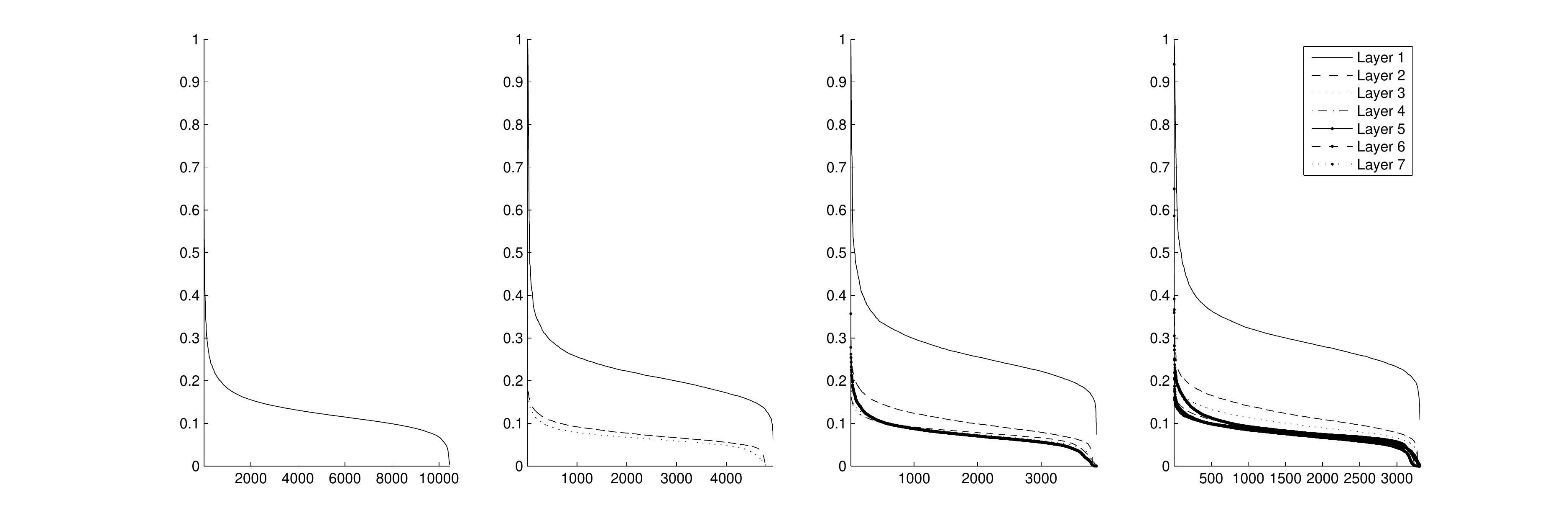}}
\centerline{\includegraphics[width=2.25\columnwidth]{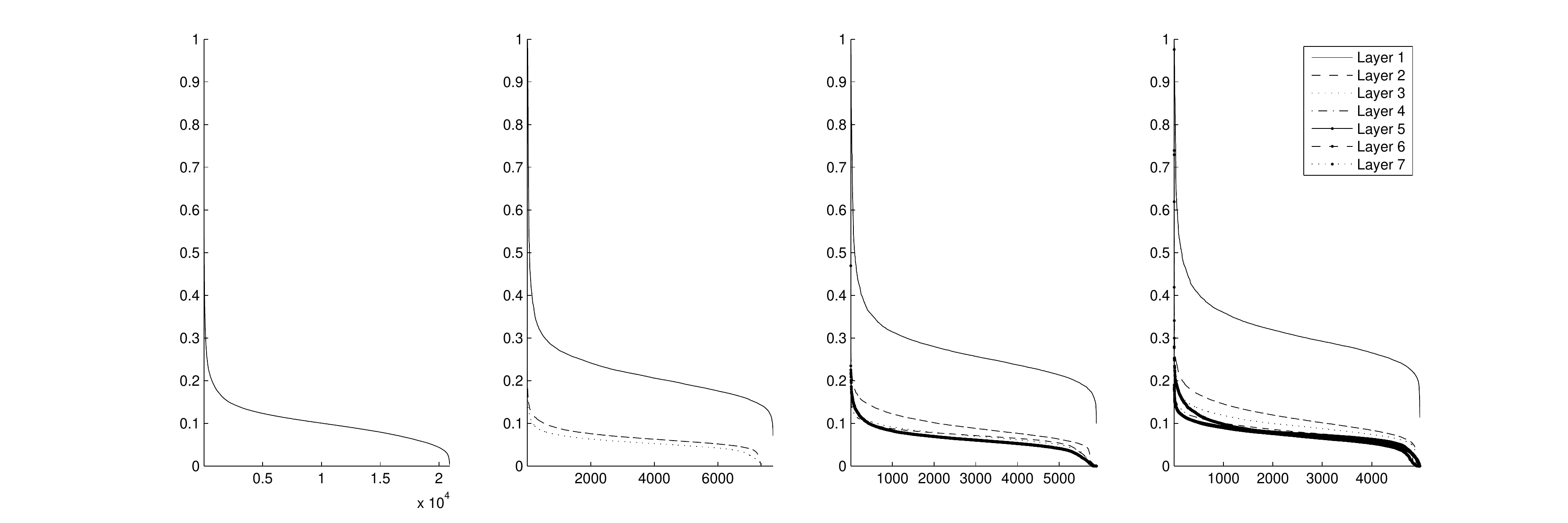}}
\caption{ Empirical activation probability of hidden units in each
  hidden layer layer of DNNs with varying numbers of hidden
  layers. Each row contains DNNs of 36M (top), 100M (middle), and 200M
  total parameters (bottom). From left to right, each sub-figure shows
  a DNN with 1, 3, 5, and 7 hidden layers.  Hidden units (x axis) are
  sorted by their probability of activation. We consider any positive
  value as active ($h(x)>0$). }
\label{fig:scree_depth}
\end{center}
\end{figure*}

Given the task performance differences observed as a function of DNN
depth for a fixed number of total DNN parameters, we also compare
scree plots as a function of DNN depth to better understand their
coding properties. Figure~\ref{fig:scree_depth} shows scree plots for
DNNs with 1, 3, 5, and 7 hidden layers for DNNs of total size 36M,
100M, and 200M. We observe a general trend of average activation
probability decreasing in subsequent hidden layers of DNNs at each
size. This is not true, however, for models with 7 hidden layers,
which have slightly less sparse activations on average in layers 6 and
7 as compared to layer 5. As we compare models across total model size we find that larger models are more sparse than smaller models. Larger models also tend to be slightly more dispersed on average compared with smaller models.

\subsection{Code Length}
\begin{figure*}[tb]
\begin{center}
\centerline{\includegraphics[width=2.25\columnwidth]{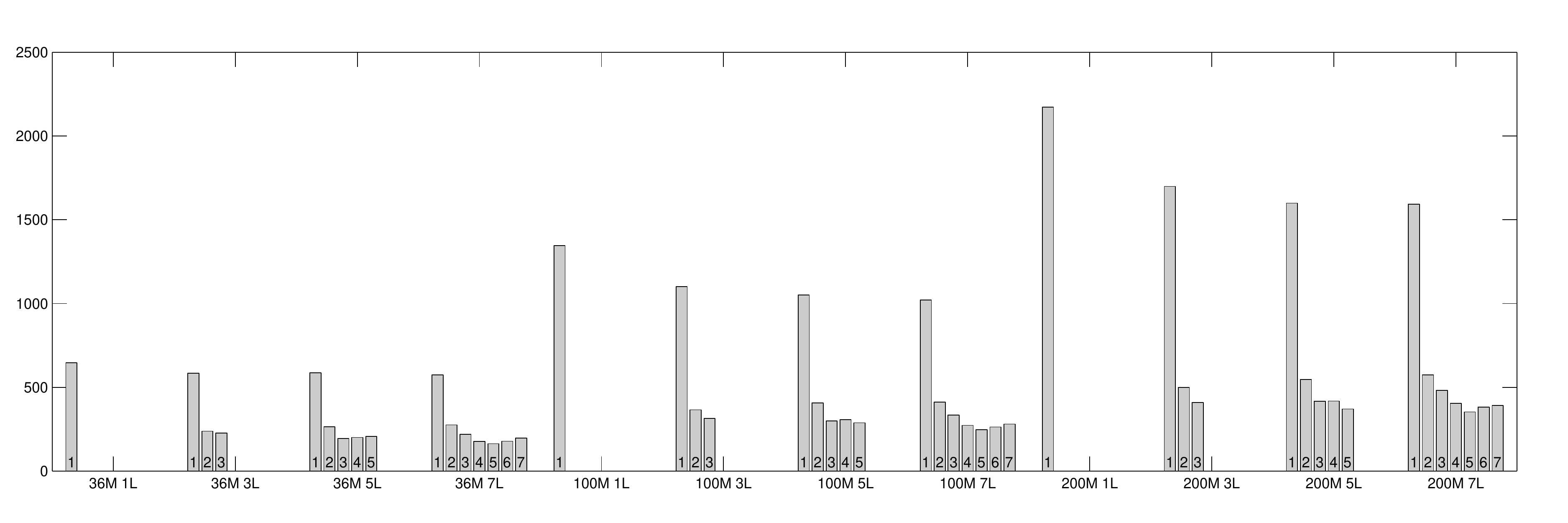}}
\caption{ Effective code length for each hidden layer in DNNs with varying total size and depth. 
  We compute the number on non-zero hidden unit
  activations for a given input, and then average over a large sample
  of inputs. Plots show the average number of units active in each
  hidden layer of DNNs of varying depth and total size. Within each
  sub-plot layers are ordered left to right from first to final hidden
  layer. }
\label{fig:code_length}
\end{center}
\end{figure*}

Our sparsity and dispersion metrics serve as indicators for how hidden
units within each layer behave. We now focus on \emph{code length},
which analyzes each hidden layer as a transformed representation of
the input rather than focusing on individual units with each hidden
layer. For a given input we compute the number of non-zero hidden unit
activations in a hidden layer. We can then compute the average code
length for each hidden layer over a large sample of inputs from our
dataset. Figure~\ref{fig:code_length} shows average code length for
each hidden layer of DNNs of varying depth and total size.

As we compare code length across models of varying total parameter
size, we see that larger DNNs use more hidden units per layer to
encode information at each hidden layer. This trend is especially
evident in the first hidden layer, where 100M parameter models use
nearly twice the code length as compared to 36M models. In deeper
layers, we again observe that models with more parameters have greater
code length. It is unclear to what extend the longer codes are
capturing more information about an input, which in turn should enable
greater classification accuracy, versus redundancy where
multiple hidden units encode overlapping information. 

Code length in deeper versus more shallow models of the same total
size exhibit an interesting trend. DNNs of increasing depth show a
generally decreasing or constant code length per layer, except in the
case of our 7 hidden layer DNNs. In 7 hidden layer DNNs, the deepest
models we trained, code length decreases until it reaches a minimum at
layer 5, but then increases in layers 6 and 7. This trend is evident
in models of 36M, 100M, and 200M total parameters. We note that this
trend of decreasing code length followed by increasing code length is
correlated with the lack of improvement of 7 hidden layer models as
compared to 5 hidden layer models. More experiments are needed to
establish whether code length in deeper models is more generally
correlated to diminishing task performance.

%% file: conclusion.tex
\section{Conclusion}
\label{sec:conclusion}
The multi-step process of building neural network acoustic models
comprises a large design space with a broad range of previous
work. Our work sought to address which of the most fundamental DNN
design decisions are most relevant for final ASR system
performance. We found that increasing model size and depth are simple
but effective ways to improve WER performance, but only up to a
certain point. For the Switchboard corpus, we found that
regularization can improve the performance of large DNNs which
otherwise suffer from overfitting problems. However, a much larger
gain was achieved by utilizing the combined 2,100hr training corpus as
opposed to applying regularization with less training data.

Our experiments suggest that the DNN architecture is quite competitive
with specialized architectures such as DCNNs and DLUNNs. The DNN
architecture outperformed other architecture variants in both frame
classification and final system WER. While previous work has used more
specialized features with locally connected models, we note that DNNs
enjoy the benefit of making no assumptions about input features having
meaningful time or frequency properties. This enables us to build DNNs
on whatever features we choose, rather than ensuring our features
match the assumptions of our neural network. We found that DLUNNs
performed slightly better and DCNNs, and may be an interesting
approach for specialized acoustic modeling tasks. For example, locally
untied models may work well for robust or reverberant recognition
tasks where particular frequency ranges experience interference or
distortion.

We trained DNN acoustic models with up to 400M parameters and 7 hidden
layers, comprising some of the largest models evaluated to date for
acoustic modeling. When trained with the simple NAG optimization
procedure, these large DNNs achieved clear gains on both frame
classification and WER when the training corpus was large. An analysis
of performance and coding properties revealed a fairly gradual change
in DNN properties as we move from smaller to larger models, rather
than finding some phase transition where large models begin to encode
information differently from smaller models. Overall, total network
size, not depth, was the most critical factor we found in our
experiments. Depth is certainly important with regards to having more
than one hidden layer, but differences among DNNs with multiple hidden
layers were fairly small with regards to all metrics we
evaluated. At a certain point it appears that increasing DNN
depth yields no performance gains, and may indeed start to harm
performance. When applying DNN acoustic models to new tasks it appears
sufficient to use a fixed optimization algorithm, we suggest NAG, and
cross-validate over total network size using a DNN of at least three
hidden layers, but no more than five. Based on our results,
this procedure should instantiate a reasonably strong baseline system
for further experiments, by modifying whatever components of the
acoustic model building procedure researchers choose to explore.

Finally, we note that a driving factor in the uncertainty around DNN
acoustic model research stems from training the acoustic model in
isolation from the rest of the larger ASR system. All models trained
in this paper used the cross entropy criterion, and did not perform as
well as DNNs trained with discriminative loss functions in previous
work. We hypothesize that large DNNs will become increasingly useful
as researchers invent loss functions which entrust larger components
of the ASR task to the neural network. This allows the DNN to utilize
its function fitting capacity to do more than simply map acoustic
inputs to HMM states.

We believe a better understanding of task performance and
coding properties can guide research on new, improved DNN
architectures and loss functions. We trained DNNs using
approximately 300 lines of Python code, demonstrating the feasibility
of fairly simple architectures and optimization procedures to achieve
good system performance. We hope that this serves as a reference point
to improve communication and reproducibility in the now highly active
research area of neural networks for speech and language understanding.